%% file: main.tex
\documentclass[10pt,twocolumn,letterpaper]{article}

\usepackage[pagenumbers]{misc/cvpr}

\usepackage{misc/mo}
\usepackage{inconsolata}

\makeatletter
\newcommand\footnoteref[1]{\protected@xdef\@thefnmark{\ref{#1}}\@footnotemark}
\makeatother

\newif\ifdrafting
\draftingtrue 
\draftingfalse 
\ifdrafting
    \newcommand{\todo}[1]{{\leavevmode\color[rgb]{1,0,0}[TODO: #1]}}    
    \newcommand{\diff}[2]{{\color{red}\sout{#1}} {\color{green}#2}}
    \newcommand{\missing}[1]{{\color{orange}{#1}}}
    \newcommand{\mo}[1]{{\leavevmode\color[rgb]{0.25,0.4,0.88}[\textbf{Mo}: #1]}}
    \newcommand{\vj}[1]{{\color{magenta}{\textbf{Varun: } #1}}}
    \newcommand{\km}[1]{{\color{blue}{\textbf{Kevis: } #1}}}
    \newcommand{\ds}[1]{{\leavevmode\color[rgb]{0.8,0,0.8}[Deqing: #1]}}
    
\else
    \newcommand{\todo}[1]{}
    \newcommand{\diff}[1]{}
    \newcommand{\missing}[1]{}
    \newcommand{\mo}[1]{}
    \newcommand{\vj}[1]{}
    \newcommand{\km}[1]{}
    \newcommand{\ds}[1]{}
\fi

\def\paperID{470} 
\def\confName{CVPR}
\def\confYear{2024}

\title{Probing the 3D Awareness of Visual Foundation Models}

\author{
Mohamed El Banani$^1$ \quad 
Amit Raj$^2$ \quad 
Kevis-Kokitsi Maninis$^2$ \quad
Abhishek Kar$^2$ \quad
Yuanzhen Li$^2$ \\ 
Michael Rubinstein$^2$ \quad
Deqing Sun$^2$\quad
Leonidas Guibas$^2$ \quad
Justin Johnson$^1$ \quad
Varun Jampani $^{2*}$
\\[3pt]
$^1$~University of Michigan \qquad $^2$~Google Research
\vspace*{-10pt}
}

\begin{document}
\maketitle

\global\csname @topnum\endcsname 0
\global\csname @botnum\endcsname 0

\begin{NoHyper}
  \let\thefootnote\relax\footnotetext{* Current affiliation is Stability AI.}
\end{NoHyper}

\begin{abstract}
    \input{sections/s0_abstract}

\end{abstract}

\input{sections/f1_teaser}

\section{Introduction}
\label{sec:introducion}
\input{sections/s1_intro}

\section{3D Aware Visual Representations}
\label{sec:evaluate_3d}
\input{sections/s2_3d_understanding}

\section{Experimental Setup}
\label{sec:experiments}
\input{sections/s3_experiments}

\subsection{Single Image Surface Reconstruction}
\label{subsec:singleview_experiments}
\input{sections/s3a_singleimage}

\subsection{Multiview Consistency}
\label{subsec:mutliview_experiments}
\input{sections/s3b_multiview}

\subsection{Analysis}
\label{subsec:analysis}

\input{sections/s3c_analysis}

\section{Related Work}
\label{sec:related}
\input{sections/s4_related}

\section{Discussion}
\label{sec:discussion}
\input{sections/s5_conclusion}

{
\small
\vspace{0.2cm}
\noindent 
\textbf{Acknowledgments:}
This work was done during an intership with the VisCAM team at Google Research.
We thank Prafull Sharma, Shivam Duggal, Karan Desai, Junhwa Hur, and Charles Herrmann for many helpful discussions.
We also thank Alyosha Efros, David Fouhey, Stella Yu, and Andrew Owens for their feedback. 
} 

{
\small
\bibliographystyle{misc/ieeenat_fullname}
\bibliography{references}
}

\appendix
\renewcommand{\thesection}{\Alph{section}} 

\maketitlesupplementary


\section{Additional Experimental Details}
\label{app:experimental_details}

We provide a high-level summary of the experimental setup in the main body of the paper and omitted details to enhance readability. In this section, we provide a more extensive coverage of our experimental setup, as well as explain the rationale behind some of our design choices. 
We will also release the code at \texttt{\hyperlink{https://github.com/mbanani/probe3d}{https://github.com/mbanani/probe3d}} to allow others to replicate and expand on our analysis. 

\subsection{Visual Foundation Models}
\label{app:models}
\input{supplemental/a_models}

\subsection{Evaluation Datasets}
\label{app:datasets}
\input{supplemental/a_datasets}

\subsection{Evaluation Tasks}
\label{app:tasks}
\input{supplemental/a_tasks}

\subsection{Performance Correlation}
\label{app:correlations}
\input{supplemental/a_correlation}

\section{Additional Results}
\label{app:complete_results}
\input{supplemental/app_complete_results}

\section{Limitations}
\label{app:limitations}
\input{supplemental/app_limitations}

\clearpage
\input{supplemental/tab_depth_all}

\input{supplemental/tab_snorm_all}

\input{supplemental/tab_corr_scannet}

\input{supplemental/tab_corr_navi}

\input{supplemental/tab_corr_spair}

\clearpage

\end{document}

%% file: sections/s0_abstract.tex
Recent advances in large-scale pretraining have yielded visual foundation models with strong capabilities. 
Not only can recent models generalize to arbitrary images for their training task, 
their intermediate representations are useful for other visual tasks such as detection and segmentation.
Given that such models can classify, delineate, and localize objects in 2D, we ask whether they also represent their 3D structure?
In this work, we analyze the 3D awareness of visual foundation models.
We posit that 3D awareness implies that representations (1) encode the 3D structure of the scene and (2) consistently represent the surface across views.
We conduct a series of experiments using task-specific probes and zero-shot inference procedures on frozen features. 
Our experiments reveal several limitations of the current models. 
Our code and analysis can be found at \texttt{\hyperlink{https://github.com/mbanani/probe3d}{https://github.com/mbanani/probe3d}}.

%% file: sections/f1_teaser.tex
\begin{figure}
    \centering
    \includegraphics[width=\linewidth]{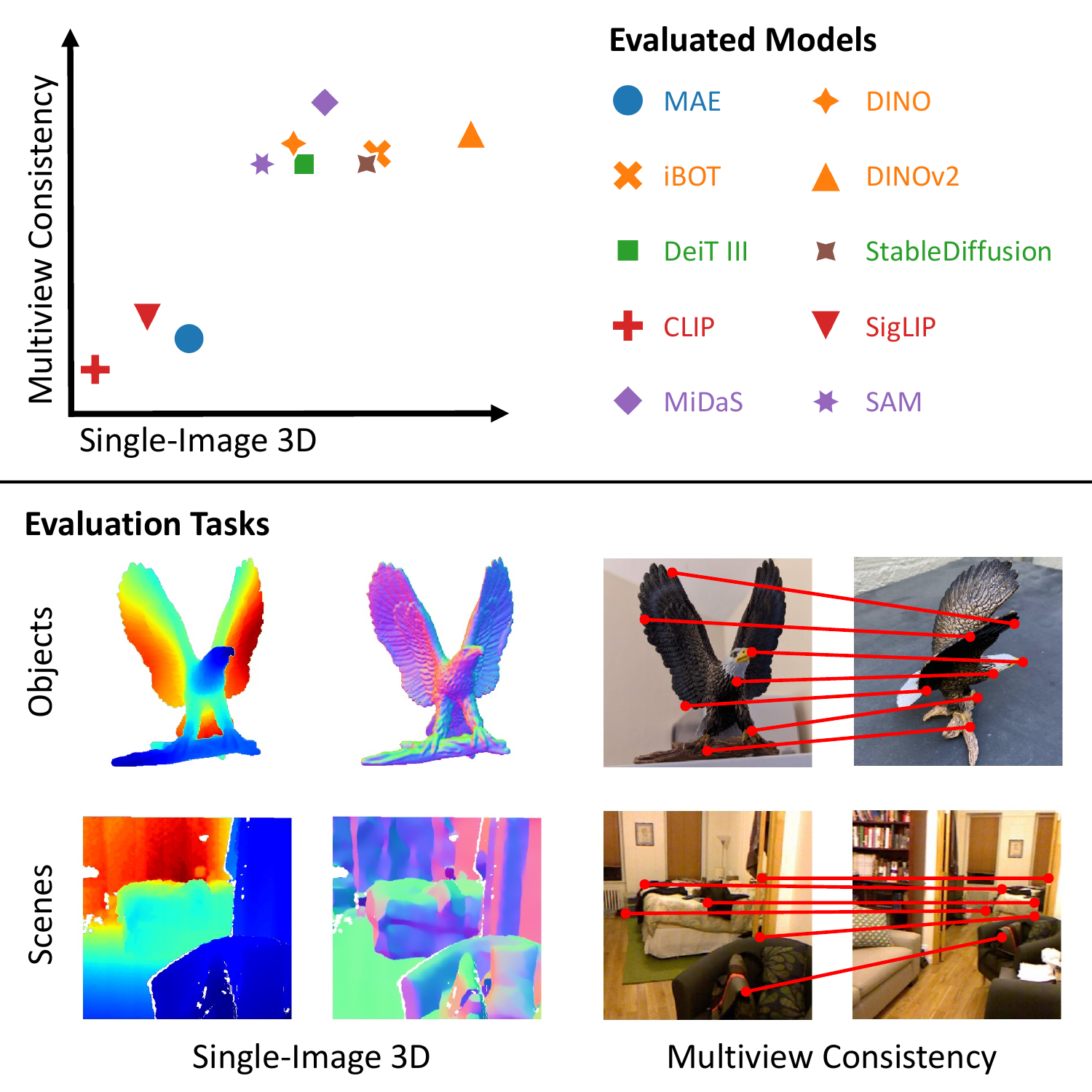}
    \caption{\textbf{Are current visual foundation models 3D aware?} 
    We probe the 3D awareness of the learned representations by evaluating their ability to encode the 3D structure of the visible surface and their consistency across views. 
    }
    \label{fig:teaser}

\end{figure}

%% file: sections/s1_intro.tex
Large-scale pretraining on image datasets has yielded visual foundation models with impressive generalization capabilities. 
Such models can classify~\cite{radford2021learning,li2021align}, segment~\cite{kirillov2023sam}, and generate~\cite{rombach2022high,chang2023muse,saharia2022photorealistic} arbitrary images. 
Furthermore, the dense representations learned by such models extend beyond their training tasks and exhibit strong zero-shot capabilities in other tasks such as segmentation~\cite{xu2023odise,melas2022deep} and part discovery~\cite{amir2021deep,gupta2023asic}.
This suggests that models are learning strong image representations, but how well do they represent the 3D world that images depict?

Recent work suggests that visual foundation models are useful for some 3D tasks despite being trained with 2D data.
For instance, models implicitly represent depth and surface normals when generating images of scenes and faces~\cite{chen2023beyond,bhattad2023stylegan}.
Furthermore, the intermediate representations of self-supervised and generative models can be used to estimate semantic correspondence~\cite{zhang2023tale,hedlin2023unsupervised,tang2023dift,amir2021deep,gupta2023asic} and object pose~\cite{goodwin2022zero}. 
However, when reconstructing 3D objects, they generate artifacts that suggest a lack of 3D consistency~\cite{liu2023zero}; \eg, animals with multiple faces.
Therefore, it remains unclear how those modes represent or understand the 3D world. 

The goal of this paper is to study the 3D awareness of visual foundation models. 
Previous benchmarks evaluate visual models on semantic tasks~\cite{walmer2023teaching,goyal2019scaling,goldblum2023battle}, but their 3D understanding remains understudied. 
Representations can vary from having no 3D awareness (\eg, class label or bag of words) to accurately representing the 3D geometry of the scene (\eg, 3D surface map or mesh). 
We posit that 3D awareness can be evaluated through two distinct capabilities: \textit{single-view surface reconstruction} and \textit{multiview consistency}. 
If a model is 3D aware, we expect that its representations would encode the geometry of the surfaces visible in the image; \ie, how far is each point in the scene? what is the orientation of the surface?
Moreover, the representations should be consistent for different views of the scene; allowing them to establish accurate correspondence.

To this end, we conduct an empirical analysis of the 3D awareness of visual foundation models. 
Our analysis considers a range of large-scale pretrained models that have exhibited strong generalization, regardless of their pretraining objective.
We evaluate the models on their ability to estimate 3D quantities that match the aforementioned capabilities: depth, surface normals, and 3D correspondence. 
Furthermore, we evaluate those capabilities at both the scene level~\cite{silberman2012indoor,dai2017scannet} and for individual objects~\cite{jampani2023navi,min2019spair} to provide further differentiation. 
We show the models and tasks considered in \cref{fig:teaser}.
Since we are interested in what the models represent, we probe the frozen representations through task-specific probes or zero-shot inference methods. This allows us to evaluate the models' representations, rather than the transferability of their pretrained weights.

Our experiments reveal a large variation in the 3D awareness of the models. We present the aggregated ratings (higher is better) of different models in single-image and multiview tasks in \cref{fig:teaser}.
We find that recent self-supervised models such as DINOv2~\cite{oquab2023dinov2} learn representations that encode depth and surface normals, with StableDiffusion~\cite{rombach2022high} being a close second. 
Meanwhile, the training in vision language for models such as CLIP~\cite{radford2021learning} exhibits very poor performance despite its impressive semantic generalization capabilities. 
At their best, some of the probes achieve a performance close to that of state-of-the-art models despite being pretrained with a very different objective.
Meanwhile, we find that the models struggle with multiview consistency. 
Although most models can accurately match objects and scenes with small viewpoint changes, they perform very poorly at large viewpoint variations. 
Our analysis further suggests that consistency across images is semantic in nature; \ie, models accurately match semantic parts but struggle to incorporate the global object pose.  
We hope that our findings will spark more interest in better understanding the 3D awareness of vision foundation models and contribute to more comprehensive benchmarks of visual representation learning approaches.

%% file: sections/s2_3d_understanding.tex
We first discuss what we mean by 3D aware visual representations.
When we view a scene, we seem to effortlessly understand its 3D structure despite only seeing its 2D projection. 
Research in developmental psychology and psychophysics suggests that our perception encodes surface properties such as depth and orientation~\cite{spelke2010beyond,koenderink1992surface}. 
Research on mental imagery has suggested that our internal representations of objects encode their 3D shape and are subject to 3D constraints~\cite{shepard1971mental}. 
Inspired by this work, we posit that 3D aware representations encode basic 3D properties of the surface as distances and orintations. 
Beyond a single image, 3D aware representations are consistent across views of the same object or scene, as they are projections of the same underlying 3D geometry. 

Representations in computer vision have varied a lot in how well they represented the 3D shapes of objects. 
Early representations such as 2.5D sketches~\cite{marr1979computational} and generalized cylinders~\cite{binford1971,brooks1981symbolic} explicitly depicted the 3D geometry of the obejcts and their spatial relationships.
Recent advances have deviated from explicit modeling and instead rely on the representation of visual information as dense feature grids~\cite{he2015delving} or sets of tokens~\cite{dosovitskiy2020vit}.
While 3D awareness of early representations was obvious, it remains unclear what the learned representations encode or how 3D aware they are. 
Popular interpretability mechanisms such as GradCAM~\cite{selvaraju2017grad} are not helpful here, as they tell us which components of the image led to a specific inference, not what information was represented by the network.

We propose evaluating the 3D awareness of visual models by probing them on two capabilities: single-view 3D and multiview consistency. 
We take inspiration from the work on human perception~\cite{spelke2010beyond,shepard1970second,koenderink1979internal} and evaluate models on how well they encode basic 3D properties and how 3D consistent they are. 
For a single image, we expect a 3D aware model to accurately represent the visible surface and encode properties such as depth and surface orientation. 
When given multiple images of the same object or scene, we expect a 3D aware representation to capture the relationships between the images and provide accurate correspondence. 
Although these two capabilities are not exhaustive, they capture two fundamental aspects of 3D understanding.
Furthermore, they can be directly mapped to three well-studied problems in computer vision, namely, estimating monocular depth, surface normals, and correspondence.

%% file: sections/s3_experiments.tex
The goal of our experiments is to evaluate the 3D awareness of visual foundation models: \ie, large-scale pretrained models that are proposed as general backbones for a wide variety of downstream tasks or applications. 
Specifically, we hope to answer the following questions: 
\vspace{3pt}
\begin{enumerate}
\item Do models learn to represent the visible surface?
\item Are the representations consistent across views? 
\item How does the training objective impact 3D awareness?
\end{enumerate}
\vspace{3pt}

\lsparagraph{Models.} We primarily focus our experiments on vision transformers that were proposed as general purpose backbones or that exhibit strong generalization performance across tasks or domains. 
Moreover, we are interested in evaluating models that were trained with different supervisory signals. 
First, we consider three forms of supervision that commonly serve as pretraining tasks: classification~\cite{touvron2022deit}, language supervision~\cite{openclip,radford2021learning}, and self-supervision~\cite{caron2021emerging,oquab2023dinov2,zhou2021ibot,he2022masked}.
Recent work has also shown that text-conditioned image generation~\cite{rombach2022high} can learn strong representations and provide strong backbones for other vision tasks~\cite{zhao2023unleashing,xu2023odise,li2023diffusion}. 
We also consider two forms of dense supervision that have recently been scaled up: depth estimation~\cite{ranftl2020towards,ranftl2021dpt} and class-agnostic segmentation~\cite{kirillov2023sam}.
While there models have not been used as general purpose backbones yet, they exhibit impressive generalize to a wide range of domains and provide an interesting point of comparison. 
We present an overview of the models considered in \cref{tab:models_used}, and more details can be found in \cref{app:models}.

One challenge is how to fairly compare models that have different data and compute requirements. 
This challenge is further amplified by considering the scale used to achieve the strong performance displayed by such models. 
Furthermore, the data used to train many of these models is private~\cite{radford2021learning,oquab2023dinov2} and even replicating the data collection and curation process requires extensive resources as shown by \citet{xu2023demystifying}. 
Beyond data scale and curation, models have different data requirements that range from class labels~\cite{touvron2022deit},  captions~\cite{radford2021learning}, masks~\cite{kirillov2023sam}, or
even simple curation~\cite{oquab2023dinov2}. As a result, it is unclear which dataset would provide a fair comparison. 
We make a pragmatic choice of relying on publicly available checkpoints and selecting checkpoints of comparable architectures and training scale to provide some fair comparison. 
We provide additional comparisons in \cref{app:complete_results} and discuss the impact of this on our results in \cref{app:limitations}.

\input{sections/t1_models}

Another important question is how to evaluate those properties. 
One common approach is transfer learning, where the pretrained model is fine-tuned using task-specific supervision. 
This is often a good practical choice, as it results in strong downstream performance. 
However, it is not suitable for our analysis, as good fine-tuning performance may indicate two different things: the model has good 3D awareness or the model weights are a good initialization for other tasks~\cite{goyal2019scaling}.
Furthermore, fine-tuning specializes the models by sacrificing its generality~\cite{kumar2022fine}. 
Instead, we probe the frozen features with trainable probes and zero-shot inference methods that do not change model weights or significantly alter model capacity. 
This allows us to evaluate pretrained representations of models with the assumption that the same model may be used for a wide range of tasks.

%% file: sections/t1_models.tex
\begin{table}[t!]
  \centering
    \caption{
        \textbf{Evaluated Visual Models. } We consider a range of visual models spanning several forms of supervision. We evaluate publicly available checkpoints and choose checkpoints of comparable model and training size whenever possible.
        }
    \label{tab:models_used}
  \setlength\tabcolsep{5pt}
  \footnotesize
  \begin{tabularx}{\linewidth}{X ccc }

\toprule
\textbf{Model} & \textbf{Architecture} & \textbf{Supervision} & \textbf{Dataset} \\
\midrule
DeIT III~\cite{touvron2022deit}         & ViT-B/16      & Classification    & ImageNet-22k  \\
MAE~\cite{he2022masked}                 & ViT-B/16      & SSL               & ImageNet-1k   \\
iBOT~\cite{zhou2021ibot}                & ViT-B/16      & SSL               & ImageNet-1k   \\
DINO~\cite{caron2021emerging}           & ViT-B/16      & SSL               & ImageNet-1k  \\
DINO v2~\cite{oquab2023dinov2}          & ViT-B/14      & SSL               & LVD-142M      \\
CLIP~\cite{radford2021learning}         & ViT-B/16      & VLM               & WIT-400M      \\
SigLIP~\cite{zhai2023sigmoid}           & ViT-B/16      & VLM               & WebLI         \\
StableDiffusion~\cite{rombach2022high}  & UNet          & Generation        & LAION         \\
SAM~\cite{kirillov2023sam}              & ViT-B/16      & Segmentation      & SA-1B         \\
MiDaS~\cite{ranftl2020towards}          & ViT-L/16      & Depth             & MIX-6         \\
\bottomrule
\end{tabularx}
\end{table}

%% file: sections/s3a_singleimage.tex
\input{sections/f2_depth_qualitative}

In this section, we analyze how well the models represent the visible surface in the image. 
We consider two tasks for single-view 3D understanding: depth estimation and surface normal estimation. 
Those tasks are well established in computer vision and are commonly studied in human perception and development~\cite{spelke2010beyond}. 
Although depth and surface normals are closely related quantities, they are different prediction tasks as they rely on different visual cues, as discussed by \citet{koenderink1992surface} and \citet{fouhey2016thesis}. 
We briefly outline our evaluation setup below, and refer the reader to \cref{app:experimental_details} and our code release for more specific details. 

\lsparagraph{Monocular depth estimation} is the task of predicting the depth for each pixel in the image. 
Although early work framed the task as regression~\cite{eigen2014depth}, recent work has shown that the use of binned prediction results in better performance~\citet{bhat2021adabins}. We follow the AdaBins~\cite{bhat2021adabins} formulation and train dense probes using their proposed losses. We report the root-mean-squared prediction error for depth estimation as well as recall at different threshold rations, similar to \citet{eigen2014depth}. 

We find that estimating the depth for object-centric datasets is particularly challenged by scale ambiguity. 
While scale ambiguity affects both objects and scene, we find that models trained to estimate metric depth on objects end up focus on predicting the object's mean depth without capturing any details.
As a result, we use a scale-invariant formulation for objects by normalizing their depth between 0 and 1.

\lsparagraph{Surface normal estimation} is the task of predicting the orientation of the surface at every pixel. 
We adopt the setup of \citet{bae2021estimating}, which utilizes an uncertainty-aware angular loss. 
Similarly to \citet{fouhey20153dwithout3d}, we report the root-mean square angular prediction error as well as the precentage recall at different angular thresholds. 

\lsparagraph{Probe. }
We use a dense multiscale probe similar to the DPT decoder~\cite{ranftl2021dpt}. 
This deviates from the common choice of linear probing commonly used in self-supervised model benchmarking~\cite{kornblith2019better}.
Linear probing is useful for semantic tasks since linear separability of classes is a desired and expected property. However, it is unclear why we should require the encoding of 3D properties to be linear.
Furthermore, the model may represent such properties at different, or multiple, locations within the network.
Hence, instead of training a linear probe on a specific linear, we use a multiscale dense probe to map the features from multiple layers to either depth or surface normals. 

\lsparagraph{Optimization. }
We train the probes for 10 epochs using the AdamW~\cite{kingma2015adam,loshchilov2016sgdr} optimizer with a linear warm and cosine decay learning rate scheduler. While longer training further improves performance, trends stabilize after 5 training epochs due to the relatively small capacity of the probe. 

\paragraph{Datasets. } We evaluate the performance on both scenes and objects.
We use the NYUv2 dataset~\cite{silberman2012indoor} to evaluate scene-level performance as it is a common benchmark for indoor scene understanding. 
We evaluate object-level performance using the NAVI dataset~\cite{jampani2023navi} which includes a set of object instances in a wide range of scenes and orientations. Both datasets provide aligned depth maps. For surface normals, we use the annotations generated by \citet{ladicky2014discriminatively} and generate the surface normal annotations for NAVI.

\lsparagraph{Results. }
We evaluate all models and report the performance in \cref{app:complete_results} due to space limitations. We focus here on qualitative results and performance trends, and analyze them through a series of questions:

\lsparagraph{Do models learn to represent depth? }
We observe that the ability of the models to encode depth is highly variable.
This can be clearly seen in \cref{fig:depth_qualitative} where DINOv2 and StableDiffusion predict accurate and detailed depth maps that capture the cow's ear and chair legs, while CLIP and MAE generate blurry and inaccurate estimates. 
It is worth noting that the models compared are all highly performant models that are often used within as backbones for downstream taks. The disparity seen highlights the importance of considering a wider range of tasks for benchmarking such models, as well as the utility of 3D awareness as a domain for such benchmarking.

\input{sections/f4_snorm_qualitative}

\lsparagraph{Do models learn to represent surface normals? }
Surface normal probe results reveal similar trends to depth estimation, with some models achieving very high performance, while others struggle to capture any information beyond the coarse priors such as ``floor pixels point up.'' 
The reliance on priors becomes more clear when comparing predictions for objects and scenes since objects have fewer priors due to the large pose variation. 
This is useful when analyzing the qualitative results of CLIP, which may appear blurry but correct for scenes, but are clearly inaccurate for objects. 
However, the best-performing model, DINOv2, achieves an impressive performance that is competitive with state-of-the-art models.

\lsparagraph{How is performance correlated across both tasks? }
We observe that the performance of models is strongly correlated across domains and tasks as shown in \cref{fig:singleview_correlation}. This supports our experimental design choices as it suggests that we are measuring a single capability using different methods. Furthermore, the consistent performance across indoor scenes and objects suggests that such models are learning to represent some information about the visible surface without any task-specific supervision. Although recent work has focused on the ability of generative models to learn this information~\cite{chen2023beyond,bhattad2023stylegan}, we find that it is not unique to such models trained with classification or discriminative self-supervision achieving comparable performance.

We note that, while depth and surface normal performance are well correlated at the model level, the correlation is far weaker when considering performance at the image or pixel level. We find that model performance is not consistent at the image or patch level; \eg, we find that the correlation between errors made by DINOv2 on NYU is 0.37 when aggregating at the image level, and 0.13 when considering pixel-level errors. 
Hence, while the underlying ability to represent the surface is shared, surface normals and depth estimation rely on different visual cues\cite{koenderink1996pictorial,norman2006visual, fouhey2016thesis} resulting in model errors being weakly correlated. 

\input{sections/f3_singleview_corr}

\lsparagraph{What is the impact of the training objective?}
We observe that discriminative self-supervised models perform best across both tasks and domains. 
This is surprising since it is unclear why the self-distillation and instance descrimination losses used to train such models would encourage this behavior. 
Consistent with other work~\cite{bhattad2023stylegan,chen2023beyond}, we find that StableDiffusion also captures surface properties well. 
Interestingly, models trained with dense supervision, even depth supervision, perform worse than self-supervised and text-conditioned generation, and perform on par with classification-trained models. 
Finally, language-supervised models appear to perform poorly despite their common utility as backbones for a variety of tasks. 
This could be related to previous findings that vision language models struggle with spatial relations and compositionality~\cite{lewis2022does,subramanian2022reclip,li2024localizationvssemantics}.

Overall, our experiments suggest that most visual models suggest that most visual foundation models end up learning representations that encode properties of the visual surface despite being trained with just image data.

%% file: sections/f2_depth_qualitative.tex
\begin{figure*}[t!]
    \centering
    \includegraphics[width=\linewidth]{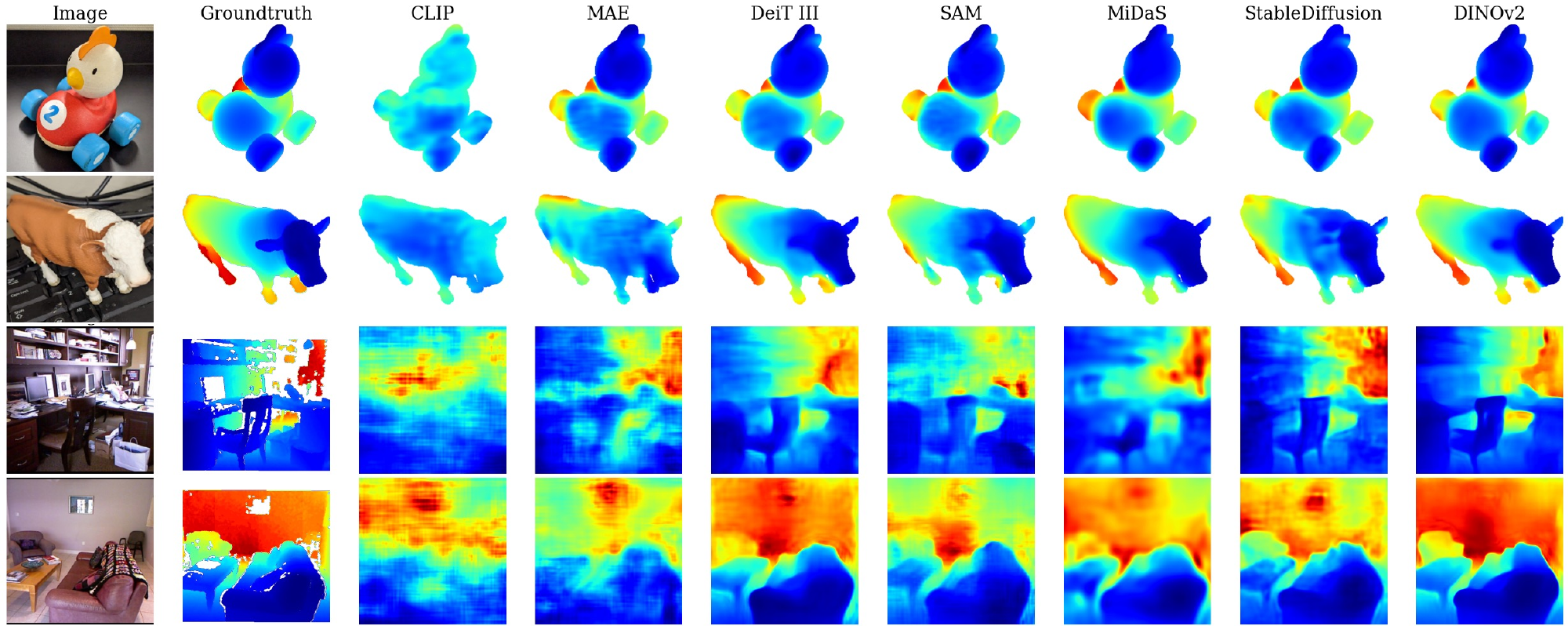}
    \caption{\textbf{Depth Estimation Results. } While pretrained representations exhibit large variation in their ability to represent depth, their performance is consistent on objects and scenes. 
    CLIP and MAE features do not encode depth and appear to instead capture rough priors such as "floor pixels are close".
    Most models appear to capture the rough structure of the scene and vary in the degree to which they capture details.
    DINOv2 performs best and accurately captures fine details; \eg, cow's ear, desk chair, and coffee table. 
    }
    \label{fig:depth_qualitative}
\end{figure*}

%% file: sections/f4_snorm_qualitative.tex
\begin{figure*}[t!]
    \centering
    \includegraphics[width=\linewidth]{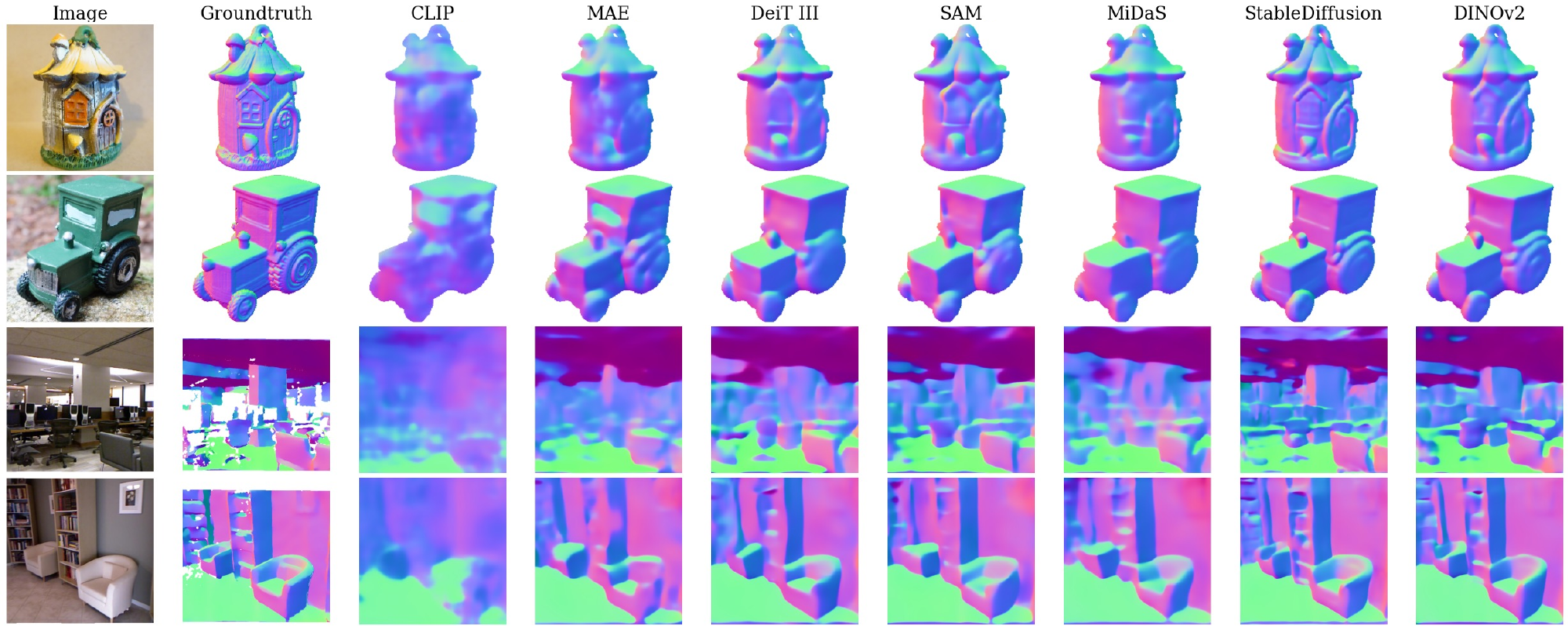}
    \caption{\textbf{Surface Normal Qualitative Examples. } 
    With the exception of CLIP, models can capture the rough orientation of object and scene surfaces; \eg, floors, walls, ceilings. 
    The main distinction seems to be in how well they capture finer details. 
    Similarly to depth results, we find that DINOv2 and StableDiffusion perform best and can capture fine details such as the edges of the toy car and the white seat.
    Surprisingly, we find that SAM's predictions are not as detailed despite its ability to predict accurate segmentation boundaries.
    }
    \label{fig:snorm_qualitative}
\end{figure*}

%% file: sections/f3_singleview_corr.tex
\begin{figure}[t!]
    \centering
    \includegraphics[width=\linewidth]{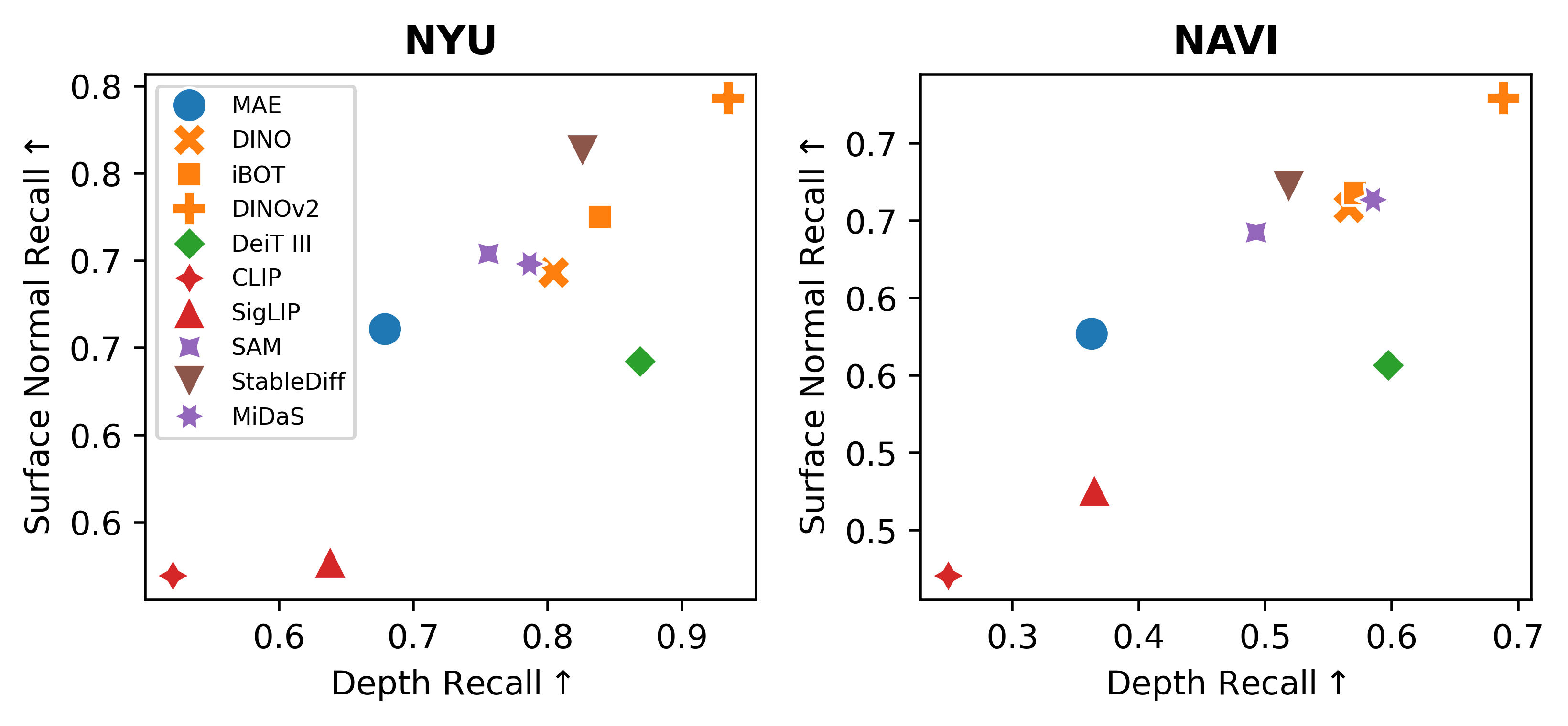}
    \caption{\textbf{Single view performance correlation. }
    Depth and surface normal performance is highly correlated across domains.
    }
    \label{fig:singleview_correlation}
\end{figure}

%% file: sections/s3b_multiview.tex
We previously evaluated the models' ability to represent the visible surfaces. 
Although this is important for 3D understanding, the evaluation is limited to a single image. 
As discussed previously, 3D awareness also implies consistency of representations across multiple views. 
We evaluate this using correspondence estimation, where the goal is to identify image patches across views that depict the same 3D point. 
This capability is important because it would allow the model to correctly aggregate information across views, which is central to reconstruction and localization pipelines.

\lsparagraph{Geometric correspondence estimation. } Given two views of the same object or scene, identify pixels across views that depict the same point in 3D space.
Rather than training a probe, we directly compute correspondence between the dense feature maps extracted from each image as this allows us to directly evaluate the consistency of the representations. 
This inference procedure is derived from keypoint-free correspondence estimation pipelines~\cite{sun2021loftr,elbanani2021unsupervisedrr,elbanani2021byoc} and is similar to recent approaches to assess feature quality~\cite{zhang2023tale,amir2021deep,tang2023dift}

\input{sections/f5_correspondence_qualitative}

\llsparagraph{Datasets.} We consider both scenes and objects. For scenes, we evaluate our model on the Paired ScanNet~\cite{dai2017scannet} split proposed by \citet{sarlin2020superglue}.  For objects, we sample view pairs from the NAVI wild set which depict the same object instances in different environments. We sample views that have a maximum rotation of 120 degrees to ensure that there exists a mutually visible surface. 
We also evaluate performance on the SPair dataset~\cite{min2019spair} which provides keypoint-labeled images allowing us to analyze the models' performance on a closely related task: semantic correspondence estimation.

\llsparagraph{Evaluation. } 
We report the correspondence recall; \ie, the percentage of correspondence that falls within some defined distance.
Correspondence error is often computed in pixels to account for the large variation in depth; \eg, a prediction off by 1 pixel can be a few millimeters on a near-by surface or several meters for outdoor scenes. 
This choice is less suitable for objects, since they do not have the same large variation depth.
Object can also suffer from self-occlusion and repeated parts, which makes a pixel-wise threshold potentially errenous. Therefore, we use a metric threshold for objects. 
Since layer selection can greatly affect performance~\cite{walmer2023teaching}, we evaluated model performance at four different intermediate points. 
Finally, we find that model performance varies greatly depending on the viewpoint differnce between the view pairs, as we discuss next. As a result, we bin the performance depending on the magnitude of the transformation between the view pairs. 
For more details on the evaluation setup, we refer the reader to \cref{app:experimental_details}.

We evaluate all models on the three datasets and report the results in \cref{app:complete_results}. We present qualitative results and performance trends in \cref{fig:corr_qualitative} and \cref{fig:multiview_corrs}.

\input{sections/f6_multiview_corr}

\lsparagraph{Are the representations 3D consistent?}
While models can estimate accurate correspondence between objects for small viewpoint changes, the performance quickly deteriorates for larger viewpoint changes, as seen in \cref{fig:multiview_corrs}. 
Although we expect performance to be lower for larger viewpoint changes as they are more difficult, the rate of deterioration is interesting.
Specifically, StableDiffusion and SAM experience very sharp drops from being among the top performers for the smallest viewpoint changes to being the worst models for the larger viewpoint changes.
This can be clearly seen in \cref{fig:corr_qualitative} where both models predict accurate dense correspondence for the dinosaur in the top row, where the viewpoint variation is minimal, but perform very poorly for the rotated eagle views.
This rapid deterioration is not universal, as shown by the wide baseline performance of DINOv2 and DeiT. 

We observe similar trends for indoor scenes where the models predict accurate correspondence when viewing the scene from a very similar vantage point, but struggle with even small viewpoint changes as seen in the last two rows of \cref{fig:corr_qualitative}. 
Although DINOv2 performs better than the other models, the absolute performances for all models are very low for wide baseline correspondence estimation. In general, our results suggest that current models are not 3D consistent despite encoding surface properties as shown in \cref{subsec:singleview_experiments}.

\input{sections/f8_spair}

\lsparagraph{Semantic vs. Geometric Correspondence. } 
Recent work has shown that self-supervised and generative models excel at estimating semantic correspondence~\cite{amir2021deep,zhang2023tale,tang2023dift}. 
Semantic correspondence~\cite{berg2005shape} generalizes the correspondence problem from matching the same points across views of the same object to matching similar semantic parts across different instances of the same class; \eg, matching a dog's left ear in images of two different dogs. 
At first glance, this seems to contradict our results, since semantic correspondence appears to capture both 3D structure and semantics.

Semantic correspondence is commonly evaluated using keypoint recall.
This evaluation makes the model's performance succeptible to semantic biases and priors in the data. 
Keypoints are often selected to be unique and easily identifiable; \eg, beaks and tails. 
Although some keypoints (\eg, eyes and knees) are repeated, they often appear in consistent spatial arrangements due to photographer bias.  

We illustrate the disparity between semantic and geometric correspondence by evaluating StableDiffusion on SPair-71k chairs in \cref{fig:spair_confusion}.
We evaluate performance using keypoint confusion rather than recall. 
We do this by matching the closest keypoint to the predicted correspondence location and plotting the confusion matrix. This is only computed for keypoints with a true match. 
While StableDiffusion estimates accurate correspondence for small viewpoint changes, it exhibits interesting error patterns for large viewpoint changes. 
Errors seem restricted to semantically related classes (\eg, seat corners, and chair legs). 
Furthermore, the qualitative results suggest that the representation captures a combination of semantics and 2D location:
\ie, the chair leg on the right. 
We suspect that this observation is related to the Janus problem observed in diffusion-based 3D reconstruction, since the same ear can be repurposed for two different faces.

%% file: sections/f5_correspondence_qualitative.tex
\begin{figure*}
    \centering
    \includegraphics[width=\linewidth]{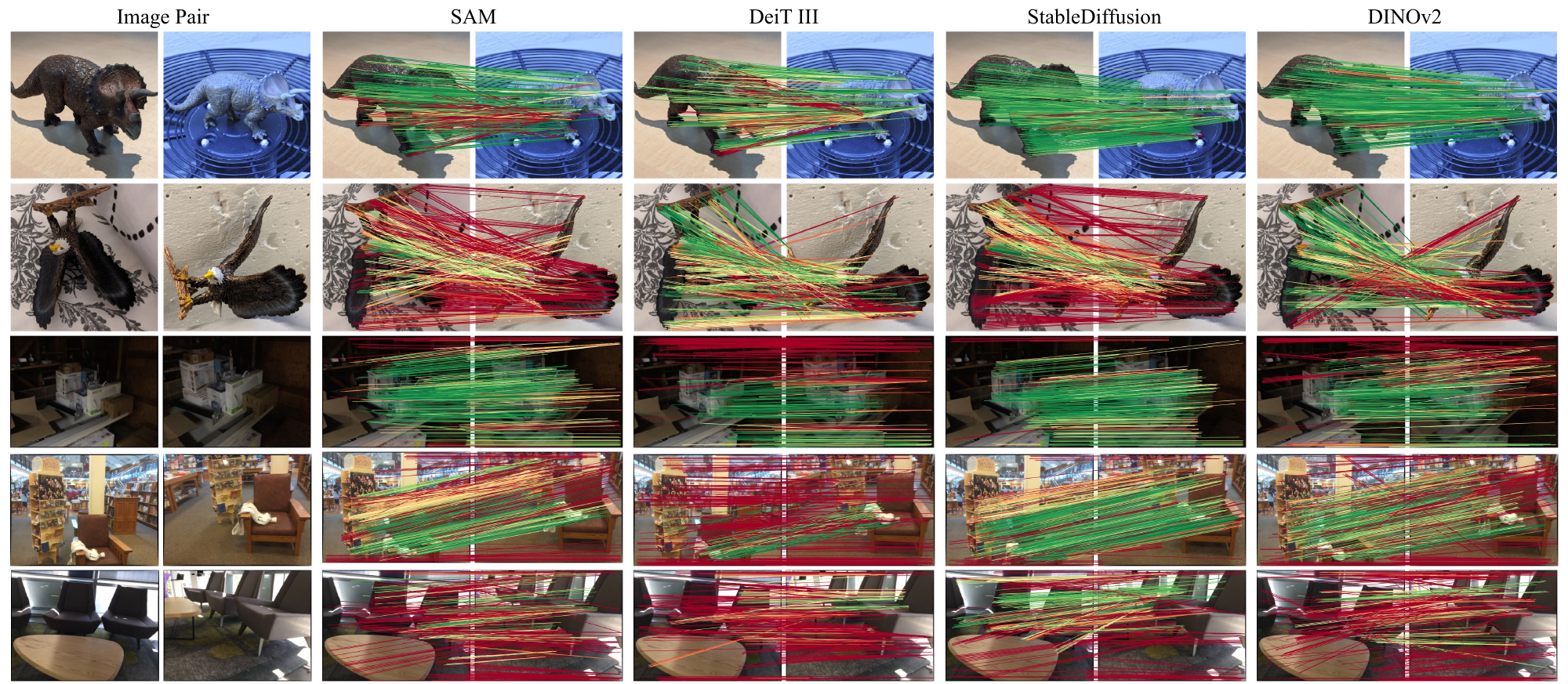}
    \caption{\textbf{Correspondence Estimation Qualitative Results. }
    We observe that models can estimate accurate correspondence for small viewpoint changes, but struggle with large viewpoint changes. 
    This is true even if the change is an in-plane rotation as shown with the eagle. 
    This pattern is consistent for both objects and scenes, although performance is not well correlated: SAM and StableDiffusion perform better for scenes, while DeiT and DINOv2 are more consistent for objects. Correspondence color-coded for accuracy.
    }
    \label{fig:corr_qualitative}
\end{figure*}

%% file: sections/f6_multiview_corr.tex
\begin{figure}
    \centering
    \includegraphics[width=\linewidth]{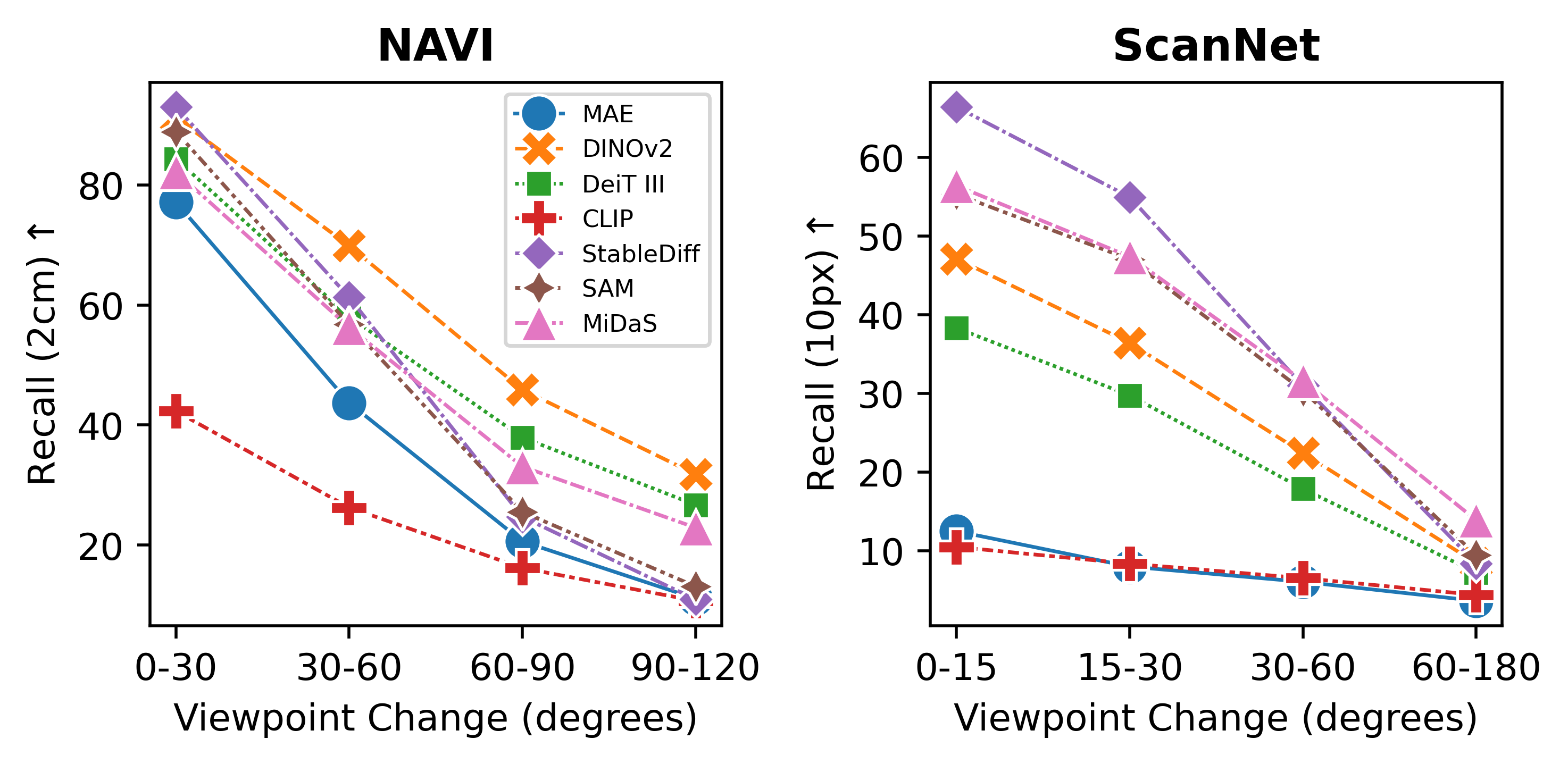}
    \caption{While all models experience performance drops with larger viewpoint changes, some experience sharper drops suggesting a lack of 3D awareness.}
    \label{fig:multiview_corrs}
\end{figure}

%% file: sections/f8_spair.tex
\begin{figure}
    \centering
    \includegraphics[width=\linewidth]{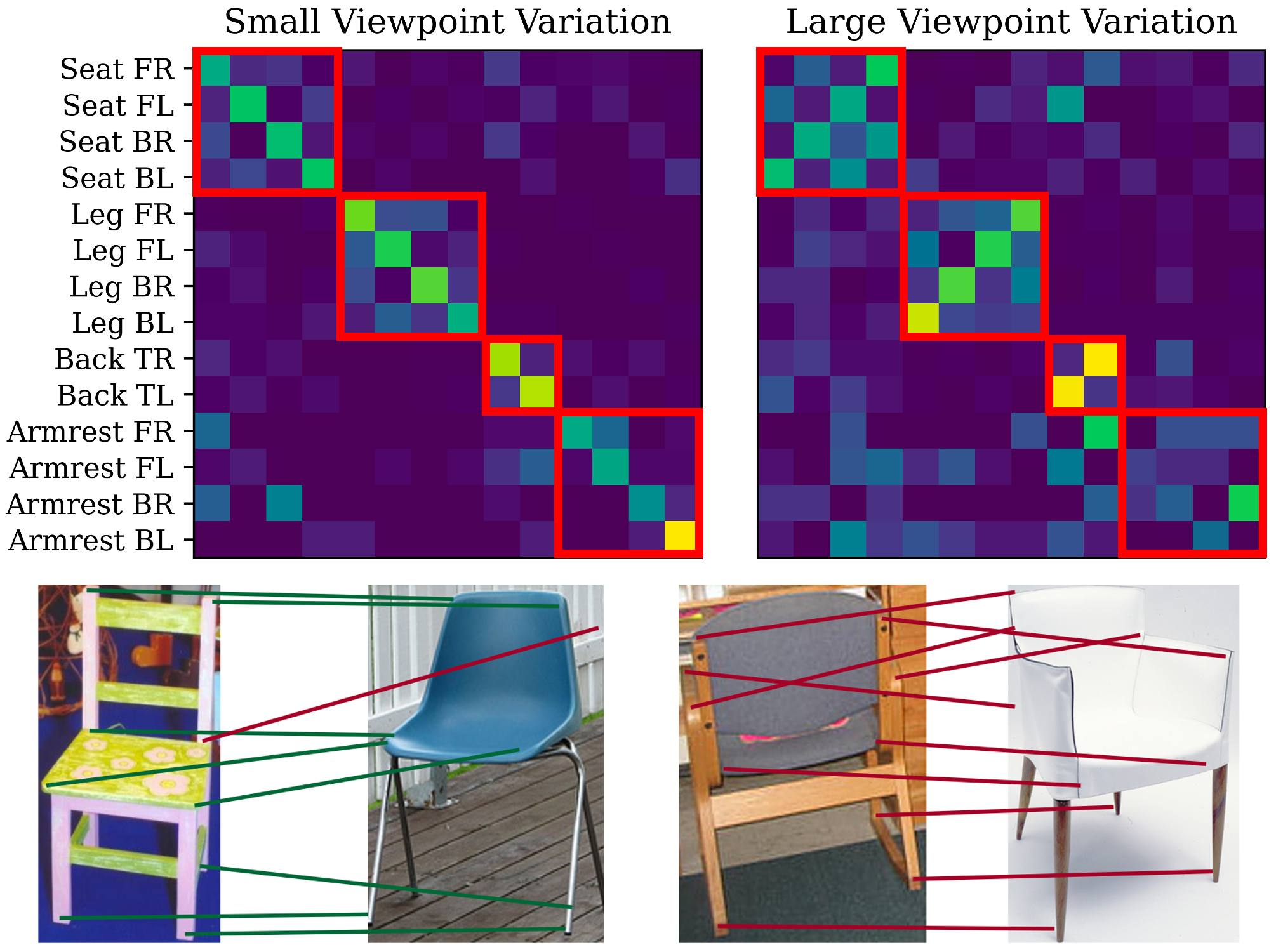}
    \caption{
    \textbf{Semantic Correspondence. }
    StableDiffusion represents semantics well, but lack 3D consistency.
    This results in accurate correspondence for objects viewed from similar angles and systematic errors when viewing objects from different viewpoints.
    \label{fig:spair_confusion}
    }
\end{figure}

%% file: sections/s3c_analysis.tex
\input{sections/f9_all_tasks_corr}

One important question is how correlated are different tasks; \ie, if a model's representations accurately represent depth, how likely is it that they are also useful for correspondence? To address this question, we compute the correlations between the models' aggregated performance across multiple tasks. We are particularly interested in understanding the relationship between training objectives and 3D awareness. We note that while we highlighted specific models in our analysis, we evaluated a much larger set of model variants and computed the cross-task performance correlations on the full set. See \cref{app:complete_results} for the complete set of results.

We compute the Pearson correlation between all pairs of tasks as shown in \cref{fig:all_task_corrs}.  For single-view 3D, we report recall for depth and surface normal estimation on objects and scene. We also report recall for correspondence estimation and separate the performance based on the amount of viewpoint variation by considering the smallest and largest viewpoint bins for NAVI and ScanNet. Finally, we also report the aggregated performance for semantic correspondence estimation.

We find that performance on all single view tasks is strongly correlated with correlation coefficients larger than 0.82. 
On the other hand, the correlation across multiview tasks is much lower, as shown by the values on the bottom right corner of the correlation matrix. 
Interestingly, semantic correspondence performance is more strongly correlated with single-view tasks than it is with multiview tasks despite having a similar evaluation procedure to the latter. This further supports our claim that semantic correspondence is not a good measure of 3D consistency.

%% file: sections/f9_all_tasks_corr.tex
\begin{figure}
    \centering
    \includegraphics[width=\linewidth]{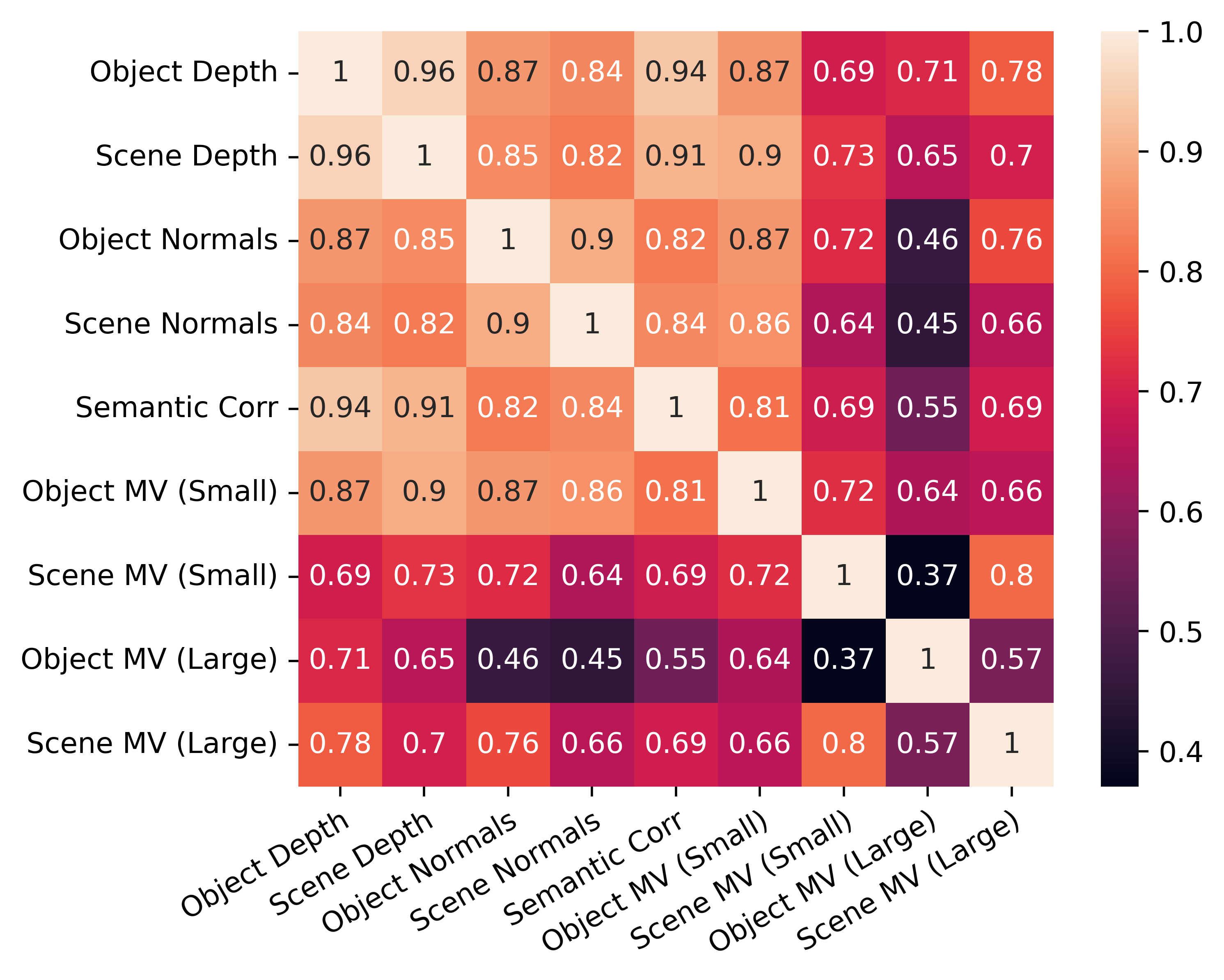}
    \caption{\textbf{Cross-task performance correlation. }
    Performance on single view tasks is strongly correlated with itself as well as semantic correspondence, but we see a drop in correlation performance of scene-level correspondence estimation and correspodence estimation with large viewpoint variation.}
    \label{fig:all_task_corrs}
\end{figure}

%% file: sections/s4_related.tex
Our work is broadly related to other efforts to understand the representations learned by vision models and to use them for 3D vision tasks.  Since the recent revival of deep learning, there has been a lot of work on understanding how and what these models learn with a focus on classification models.
Early work focused on analyzing what those models could be used for~\cite{chatfield2014return,goyal2019scaling,kornblith2019better} and providing some interpretability into what they were learning~\cite{selvaraju2017grad}. 
Our work is inspired by recent efforts to benchmark the semantic and localization capabilities of visual backbones~\cite{goyal2019scaling,walmer2023teaching,goldblum2023battle,lewis2022does,thrush2022winoground,li2024localizationvssemantics} which we try to extend towards 3D awareness. 

Recent work has attempted to evaluate the 3D understanding of vision models. 
One line of work has explored how well generative models capture single image geometry~\cite{sarkar2023geometry,du2023generative,bhattad2023stylegan,chen2023beyond}.
Although this line of work shared our goals, their probing techniques are often specific to generative models, making it difficult to extend to other visual models.
More closely related to our analysis is the recent work of \citet{zhan2023does}, who proposed analyzing the 3D understanding of StableDiffusion through a series of binary classification tasks. Instead, we focus on dense probing tasks and multiview consistency, as they are less susceptible to semantic priors, which can confound 3D undersanding, as shown by \citet{tatarchenko2019single}. Furthermore, we explore multiview consistency as another facet of 3D awareness. 

Another line of work has focus on using large-scale models for 3D tasks. 
One line of work extracts features from models for correspondence estimation~\cite{amir2021deep,oquab2023dinov2,tang2023dift,zhang2023tale,luo2023dhf,hedlin2023unsupervised} and pose estimation~\cite{goodwin2022zero,zhang2023selfsupervised}. 
Others have shown how these models could be fine-tuned for accurate depth estimation with \cite{zhao2023unleashing,ke2023repurposing} achieving state-of-the-art performance by fine-tuning StableDiffusion.  Another line of work combines image generation with 3D representations for text- or image-conditioned 3D reconstruction~\cite{wang2023score,poole2022dreamfusion,xu2023neurallift}.
While those methods generate impressive 3D shapes, it has been observed that their generations are not 3D consistent and can generate animals with multiple heads (the Janus problem).  
Recent efforts have shown that fine-tuning with 3D data can improve generation quality~\cite{kim2022dag,raj2023dreambooth3d,liu2023zero,qian2023magic123,shi2023mvdream}.
We are inspired by this line of work, but note that it differs in objective from our analysis, as we are interested in understanding 3D awareness in models trained \textit{without} 3D supervision.

%% file: sections/s5_conclusion.tex
This paper presents an exploratory study of the 3D awareness of visual models; \ie, how well do the representations capture the 3D-ness of the scenes and objects? 
We posit that 3D awareness implies representations that (1) encode the geometry of the visible surface and (2) are consistent across views. 
We used trainable probes and zero-shot inference methods to evaluate the frozen features of those models. 

Our results show that visual foundation models learn representations that encode the depth and orientation of the visible surface, with vision-language models being the notable exception. 
We also find that while models can estimate accurate semantic correspondence as well as correspondence across images of a similar viewpoint, they struggle with large viewpoint changes. 
This indicates a lack of multiview consistency and suggests that models are learning representations that are view-consistent, not 3D consistent.
One possibility is that the models are learning view-dependent representations. 
This could be similar to the theories of shape perception proposed by \citet{koenderink1979internal,koenderink1976singularities}, where shape perception is achieved by a series of view-specific representations connected with an aspect graph. 
Another possibility is that current models are simply good ``image models'' and that good discriminative features are sufficient for strong 2.5D understanding. 
We hope that our findings can simulate more interest in understanding the 3D awareness of visual models and that future work can provide better answers.

Our analysis struggles with several limitations, which we discuss in more detail in \cref{app:limitations}.
First, we used pretrained checkpoints that were often trained on different datasets and with different compute scales. While this allowed us to explore a broader set of models and tasks, 
it would be useful to make more targeted and fair comparisons to better understand the impact of training signals. 
Second, we focused on minimal probing approaches to analyze the pre-trained representations.
It would be useful to explore other probing techniques, as it remains unclear what is the best way to understand the distributed representations learned by visual models. 
Finally, our analysis only explored two basic aspects of 3D understanding.
However, 3D awareness and understanding are closely related to more complex and higher-order tasks such as perceiving 3D shapes, reasoning about spatial relationships, as well as making predictions about deformation and dynamics. 

This work is only a first step towards understanding the 3D awareness of visual models. 
This is becoming more relevant, as recent image and video generation models have achieved impressive feats of photorealism and temporal consistency. 
This makes this a very exciting time to delve into understanding what those models have learned and whether or not they learned about the 3D structure of the world in the process of learning to generate it. 
We hope that our findings will stimulate more interest in understanding the 3D awareness of visual models and that future work can provide more insight into how models represent the world and the impact of the learning objectives of such representations.

%% file: supplemental/a_models.tex
Our experiments consider 26 checkpoints that span ten learning objectives that cover five different forms of supervision. 
The models were chosen with two criteria in mind:
(1) coverage of the main approaches used for large-scale training and
(2) comparable model and training scale to allow comparisons.
We only use publicly-available checkpoints to understand the 3D awareness of the models that are commonly used. 
Although we tried to find comparable models, we note that these models were trained on different data sets using different recipes, which confounds our findings, as discussed in \cref{app:limitations}. 

We discuss all the models we used, explain their learning objective, and specify the checkpoints we used. While many of our comparisons focused on the models listed in Table 1, we used all checkpoints to calculate the correlations presented in Figure 8.
We also discuss some of the additional trends that we observed in \cref{app:complete_results}.

\lsparagraph{MAE. } \citet{he2022masked} showed that training vision transformers to reconstruct images based on randomly masked inputs is an effective pretraining task.
Such models are trained with a large masking ratio; \eg, 75\% of the input image patches are masked. 
In our experiments, we use the ViT-B/16 and ViT-L/16 models trained on ImageNet-1k. 
We use the checkpoint\footnote{\url{https://huggingface.co/facebook/vit-mae-base}} 
available on the Transformers library~\cite{wolf2020transformers}. 

\lsparagraph{FCMAE. } Fully-convolutional masked autoencoders (FCMAE) are similarly trained to reconstruct images. 
However, unlike MAE, they use a ConvNeXt backbone instead of visual transformers. 
\citet{woo2023convnext} extended the previous work to convolutional architectures and proposed several architectural changes to further improve performance. 
In our experiments, we use the ConvNeXtv2 base architecture, which has a model capability comparable to the base visual transformer architectures. 
Following the analysis of \citet{goldblum2023battle}, we use the model pre-trained on ImageNet-22k to allow a more comparable training data to other models. 
We use the checkpoints available in the timm library~\cite{rw2019timm}. \footnote{\url{https://huggingface.co/timm/convnextv2_base.fcmae_ft_in22k_in1k_384}} 

\lsparagraph{DINO. }
\citet{caron2021emerging} proposed a self-distillation approach for model pretraining. 
The proposed approach trains a student network to generate features similar to a teacher network, where the teacher is an exponential moving average of the student network. 
At its core, this approach relies on instance discrimination as the model is trained to learn to generate similar embeddings for different crops of the same image instance. In our work, we evaluate the ViT-B/16 architecture trained on ImageNet-1k. 
We use the checkpoint released by the authors.\footnote{\url{https://github.com/facebookresearch/dino}}

\lsparagraph{iBOT. }
\citet{zhou2021ibot} combine ideas from DINO and MAE by training a model to reconstruct masked dense features based on a teacher network. 
iBOT uses both an image-level and a dense distillation objective. 
We analyze the ViT-B/16 and ViT-L/16 architectures trained on ImageNet-1k and ImageNet-22k.   We evaluate the checkpoints released by the authors.\footnote{\url{https://github.com/bytedance/ibot}}

\lsparagraph{DINOv2. }
\citet{oquab2023dinov2} scale up the hybrid approach proposed by \citet{zhou2021ibot} while improving the training recipe and incorporating improved losses and regularizers. 
Furthermore, the training data and recipe are both scaled up in magnitude, resulting in much better performance. 
This includes the collection of a large curated private dataset called LVD-142M which is curated through the use of the clustered features a pre-trained self-supervised model using several downstream datasets, including NYUv2~\cite{silberman2012indoor}.
Although DINOv2 was trained on ImageNet-22k, those weights are not publicly available.
We discuss the impact of these curated datasets in \cref{app:limitations}.
We also consider DINOv2 + reg which incorporates register tokens~\cite{darcet2023vision}, however, we find that it results in slightly worse performance than the classic DINOv2 model.
We evaluate the ViT-B/14 and ViT-L/14 models for DINOv2 and the ViT-B/14 DINOv2 model with registers.  We use the checkpoints released by the authors.\footnote{\url{https://github.com/facebookresearch/dinov2}}

\lsparagraph{DeiT III.}
\citet{touvron2022deit} propose an updated training recipe for supervised vision transformers that incorporates recent best practices of self-supervised learning. 
The result is a much stronger supervised transformer compared to previous training recipes.  We evaluate the ViT-B/16 and ViT-L/16 architectures trained on ImageNet-22k. 
We use the checkpoint released by the authors.\footnote{\url{https://github.com/facebookresearch/deit}}

\lsparagraph{CLIP. }
Vision and language models are trained to generate aligned feature embeddings using a contrastive objective.
The original CLIP family of models was proposed by \citet{radford2021learning} and included a wide variety of architectures in a private dataset of 400M image-text pairs called WIT.  More recently, \citet{openclip} trained several CLIP models using several architectures trained on publicly available datasets. 
We consider five CLIP checkpoints. 
First, we evaluated the ViT-B/16 and ViT-L/14 checkpoints released by OpenAI.
Second, we evaluate the ViT-B/16 checkpoint released by OpenCLIP~\cite{openclip} that was trained on LAION~\cite{schuhmann2021laion}. \footnote{\label{fn:openclip}\url{https://github.com/mlfoundations/open_clip}}
Finally, we consider two ConvNeXt-base~\cite{liu2022convnet} checkpoints trained with and without additional augmentations~\cite{steiner2021augreg} that were also released by OpenCLIP.
We use these models to try and shed some insight on why the CLIP trained model performs poorly relative to the other models.

\lsparagraph{SigLIP. }
SigLIP~\cite{zhai2023sigmoid} also learns from images and captions, but instead replaces the contrastive objective with an instance-wise sigmoid loss.
The sigmoid loss does not require the computation of all pairs across the batch, as it only relies on the image and text embedding. 
This simplifies the objective while enabling further scaling up of the batch size for training. 
The publicly available checkpoints were trained on WebLI; a private dataset. 
We use the checkpoint available through OpenCLIP~\cite{openclip}.\footnoteref{fn:openclip}

\lsparagraph{StableDiffusion. }
StableDiffusion~\cite{rombach2022high} is trained using text-conditioned image generation using a denoising objective. 
This family of models has achieved remarkable generation performance. 
In our experiments, we use the text-conditioned checkpoint of StableDiffusion v2-1.\footnote{\url{https://huggingface.co/stabilityai/stable-diffusion-2-1}}
Following previous work~\cite{zhao2023unleashing,zhang2023tale,xu2023odise}, we extract features from the decoding blocks of the UNet. 
However, we deviate from previous work in two ways.
First, some prior work~\cite{zhan2023does,tang2023dift} computes features the image with different sampled noise and then averages them. 
While this form of ensembling is unique to diffusion-based models, it is possible to compute features based on image crops and similarly average them. 
To enable fair comparison, we simply computed features once for each image and instead experimented with different noise levels, and we observed that low noise levels appear to work best. In our experiments, we set the noise level to $t{=}1$.
Second, previous work often computes features using both the image and some auxiliary information; \eg, VPD~\cite{zhao2023unleashing} uses the image class to generate prompts for feature extraction. Such auxiliary information is not used by other models, nor is it available in many settings. Instead, we use an empty vector as a prompt similar to \citet{zhan2023does}. 

\lsparagraph{MiDaS. }
MiDaS is a family of models trained on a collection of monocular depth datasets using a scale-invariant depth estimation objective~\cite{ranftl2020towards}. 
In this work, we consider MiDaS 3.0~\cite{ranftl2021dpt} DPT Large which trains a dense prediction transformer (DPT) head on top of the features extracted from a ViT-L/16. 
While newer iterations of MiDaS 3.1 and ZoeDepth~\cite{bhat2023zoedepth} include base-size transformers, we cannot use them due to their reliance on relative positional biases.
Specifically, most ViTs rely on absolute learned or heuristic positional encoding, which can be easily interpolated to handle variable image sizes with minimal performance deterioration. However, we find that interpolating relative positional biases severely deteriorates performance.
As a result, we instead use MiDaS 3.0 which used absolute positional embeddings. 
We note that the use of a larger backbone likely exaggerates the performance of MiDaS in our analysis. 
We use the checkpoint released by the authors.\footnote{\url{https://github.com/isl-org/MiDaS}}

\lsparagraph{SAM. }
\citet{kirillov2023sam} recently proposed interactive class-agnostic segmentation as a training objective to allow generalizable open-world segmentation. 
The model is trained on a novel 10M image dataset with 1 billion masks~\cite{kirillov2023sam}.
Although the SAM architecture uses a mask decoder and a prompt encoder, the features are computed by a visual transformer backbone. 
We use the ViT-B/16 and ViT-L/16 backbones from the SAM base and large models. We use the checkpoint released by the authors.\footnote{\url{https://github.com/facebookresearch/segment-anything}}

%% file: supplemental/a_datasets.tex
\lsparagraph{NAVI. }
NAVI is a dataset of objects annotated with high-quality 3D information that was proposed by \citet{jampani2023navi}.
The dataset depicts a set of 36 objects in a wide range of poses and environments. 
High-quality object meshes are aligned to each image, which provides accurate depth and pose annotation.
We extend the dataset by generating surface normal annotations for each image. 
The dataset is organized into multiview image collections which include a larger number of multiview images of the object in the same pose and scene, as well as a wild set that depicts the object in different poses and environments. 
We use the full dataset and use multiview images for training and validation, and the wild set for testing. 
Furthermore, we exclude 2 objects from the dataset as they do not have multiview and wild-set images. 
For correspondence estimation, we only use the wild set images and for each image, we sample a pair that has a relative rotation between 0 and 120 degrees. 

\lsparagraph{NYU v2. }
The NYU Depth v2 dataset is a dataset of indoor scenes proposed by \citet{silberman2012indoor}.
The dataset consists of RGB-D video collected using a Microsoft Kinect camera and includes dense annotation for both depth and semantic segmentation. 
Furthermore, \citet{ladicky2014discriminatively} provided surface normal annotations for the labeled set of 1449 images. 
We use the original train/test split for surface normal estimation. 
For depth estimation, we also included unlabeled instances, providing us with a total of 24231 images for training. 

\lsparagraph{ScanNet Pairs. }
ScanNet~\cite{dai2017scannet} is a large dataset of RGB-D videos depicting indoor scenes. 
\citet{sarlin2020superglue} extracted a small subset of 1500 image pairs as a benchmark for correspondence estimation. 
Since our correspondence estimation experiments do not require training, we use all pairs as a test set. 

\lsparagraph{SPair 71k. }
SPair-71k~\cite{min2019spair} is a dataset of image pairs that were extracted from the PASCAL datasets~\cite{pascalvoc,pascal3d}. 
The image pairs capture a set of 18 categories and depict different object instances of the same class. Furthermore, 8 of the categories depict non-rigid objects; \eg, cats, cows, humans. 
All images are annotated with class-specific keypoints, and image pairs are further annotated with amount of viewpoint variation as judged by the human annotator. 
We follow the experimental setup of \citet{zhang2023tale} of sampling an equal number of image pairs for each class, but instead use a larger number of 200 image pairs per class. We extend this setup by separating the sampled pairs based on the annotated viewpoint variation, which is a subjective measure of how much the viewpoint changed between the two instances and is annotated with 0, 1, or 2. 

%% file: supplemental/a_tasks.tex
We evaluate all models on four tasks: monocular depth estimation, surface normal estimation, semantic correspodnence estimation, and geometric correspondence estimation. 
We chose those tasks because they evaluate the model along the two dimensions of 3D awareness we are interested in: single-image 3D understanding and multiview consistency. 
While we train the models for depth and surface normal estimation, we directly evaluate the features for correspondence. 
We note that while we use the same setup for evaluating correspondence for NAVI and NYU, SPair follows a different setup due to the existence of keypoints, which we describe separately. 
We describe each of the tasks and their evaluation procedure below. 

\subsubsection{Monocular Depth }

\lsparagraph{Task Definition: } Given an image, estimate a depth value for each pixel in the image. This problem is ill-posed because it suffers from scale ambiguity; \ie, a larger object that is further away will produce the same image as a smaller object that is closer to the camera. 
Although the regularity of our environment still allows objects to learn accurate metric depth for specific image collections, such models struggle to generalize to other image collections as different camera intrinsics or image augmentations can introduce effects similar to scale ambiguity~\cite{yin2023metric3d}. 
An alternative approach is to predict depth up to scale and then scaling it appropriately.

We use metric depth estimation for NYU due to the regularity of the data and to enable direct comparison to previous work. However, we observe that scale-invariant is more appropriate for NAVI due to the larger variance in cameras as well as a relatively small depth variation in the object surface relative to how far the object is. As a result, we scale the depth for NAVI objects between 0 and 1 for a scale-invariant depth estimation task where 0 means the closest pixel to the camera and 1 means the furthest point on the object from the camera. This variation still enables models to learn accurate depth, as shown in the main paper, and allows us to use standard depth evaluation metrics. 

We use the AdaBins~\cite{bhat2021adabins} parameterization of depth estimation due to its relatively strong performance. Rather than regressing the depth values, \citet{bhat2021adabins} proposes dividing the depth range into several bins and estimating the probability of each. The final depth value is the weighted sum of the bin probabilities and the bin center values. Similar to \citet{oquab2023dinov2}, we use 256 uniformly distributed bins and only estimate the bin probabilities. We use a depth range of 0-10m for NYU and 0-1 for NAVI. 

\lsparagraph{Losses: } We use a combination of the scale-invariant sigmoid depth loss~\cite{eigen2014depth} and the gradient matching loss~\cite{li2018megadepth} similar to \citet{oquab2023dinov2}. 

\lsparagraph{Evaluation Metrics: } We follow the evaluation setup of \citet{eigen2014depth} and compute the root mean square error and the prediction accuracy at different thresholds. The accuracy, $\delta_{i}$, is computed as the number of pixels whose ratio of depth prediction to ground-truth is less than $1.25^i$: 
\begin{align}
    \delta_i(d^{pr}, d^{gt}) = \frac{1}{N} \sum_{j \in N} \text{max}(\frac{d^{pr}_j}{d^{gt}_j}, \frac{d^{gt}_j}{d^{pr}_j}) < 1.25^{i}
\end{align}
where $d^{pr}$ is predicted depth and $d^{gt}$ is ground-truth depth. 

\lsparagraph{Probe: }
We use a non-linear multiscale convolutional probe inspired by the DPT decoder~\cite{ranftl2021dpt}.
The probe takes as input multiscale features that are extracted from several stages in the network. 
Prior work has shown that vision transformer features focus on different objects at different layers~\cite{zhang2023tale,amir2021deep} and that the granularity is not consistent between models~\cite{walmer2023teaching,zhan2023does}.
Instead of using a probe at a single layer, we train a multistage probe on features extracted from several layers.
ConvNeXt architectures often group their layers into four stages. We follow this delineation and extract features after every stage. 
For ViTs, we split the layers into 4 equally sized blocks and extract features after each block; \eg, for ViT-B, this is after layers 3, 6, 9, and 12. 
For StableDiffusion, the decoding portion of the UNet consists similarly of 4 blocks. We extract the features after each of these blocks. 
Since prior work has found that the earliest-stage features are often not useful, we only train the model on the latter three stages. This is flipped for StableDiffusion (earlier three stages) as we are sampling from the decoding part of the UNet.

\lsparagraph{Optimization: }
We train models for 10 epochs with a linear warm-up of the learning rate for 1.5 epochs and cosine decay to 0. 
We use the AdamW optimizer~\cite{liu2019radam} with a linear rate of 0.001 and a weight decay of 0.01. 

\subsubsection{Surface Normals}

\lsparagraph{Task Definition: }
Given an image, our goal is to estimate the direction of the surface at every pixel. The direction is predicted as a unit norm that is orthogonal to the surface at the point. 

\lsparagraph{Training Loss: } The cosine distance is a commonly used loss due to its relative simplicity. However, \citet{bae2021estimating} observe that the model can be severely penalized by areas that are ambiguous. As a result, they propose predicting a fourth value that captures uncertainty and calibrating the loss using that value. The loss uses the estimated uncertainty of the weight loss at each pixel, while encouraging the model to minimize its uncertainty. We use the loss formulation proposed by \citet{bae2021estimating} in our experiments. 

\lsparagraph{Evaluation Metrics: }
For each pixel, we compute the error as the relative angle between the predicted and ground-truth surface normals in degrees. 
Similar to depth estimation, we compute the RMSE for each image as well as the accuracy at different thresholds. However, instead of using the ratio as done in depth, we simply compute the accuracy at different angular thresholds ($11.25^\circ$, $22.5^\circ$, $30^\circ$) similar to previous work~\cite{bae2021estimating,piccinelli2023idisc,fouhey20153dwithout3d}. 

\lsparagraph{Probe: }
We use the same probe as depth estimation with the main difference of the final layer output dimensionality being 4 instead of 256. The four values correspond to the x-, y-, and z-components of the surface normal direction the uncertainty value used in the loss computation. We normalize the 3 directional components to a unit normal. 

\lsparagraph{Optimization: }
The optimization procedure is identical to that used for depth estimation. 

\subsubsection{Geometric Correspondence}

\lsparagraph{Task Definition: }
Given two images that depict the same object or scene from different viewpoints, the goal is to identify pairs of pixels in images that depict the same 3D point in space. We consider two settings: object-centric and scene-centric. For the scene-centric evaluation, we allow correspondences to be computed for all pixels across the images. However, for object-centric evaluation, we only consider pixels that lie on the object mask. 

\lsparagraph{Inference Procedure: } 
Given two images, we first extract a feature map for each image. 
We then estimate the correspondence using the nearest neighbors in the feature space. 
This provides us with the correspondence for each pixel, many of which will be inaccurate. 
We filter the correspondences using Lowe's ratio test~\cite{lowe2004distinctive} which aims to find unique matches by discounting points that have more than one strong correspondence. 
For each point $p$, we find its first and second nearest neighbors: $q_0$ and $q_1$. We then compute the ratio $r$ as follows:
\begin{align}
    r = 1 - \frac{D(p, q_0)}{D(p, q_1)}
\end{align}
where $D(x, y)$ is the cosine distance between $x$ and $y$. We rank the correspondences using the ratio test and keep the top 1000 correspondences. 

\lsparagraph{Evaluation: }
Correspondences are evaluated based on either 2D projection error or 3D metric error. 
Given an estimated correspondence between pixel locations $p$ in image 1 and $q$ in image 2, the 2D projection distance is computed by first projecting the point $p$ into the 3D space using known depth and intrinsics and then projecting it into image 2 using known intrinsics of the camera and the relative viewpoint between the two images.
This allows us to find the actual location of the point $p$ when projected in image 2: $p'$. The 2D correspondence error can be computed as the distance between $p'$ and $q$ in the image plane. This works very well for scenes, but can be problematic for objects where points that are not invisible can still be projected into the image plane. While it is possible to omit surface points that are not visible, the approach ignores a lot of points on thin structures; \eg, points on a wire.
Instead, we can simply compute the 3D correspondence error by focusing both points $p$ and $q$ in a shared 3D space and computing the distance between them. 
We use the 2D projection error for scenes and the 3D error for objects. 

We evaluate performance using correspondene recall; \ie, the percentage of correspondences whose error is below a specific threshold. Since we are interested in the consistency of representation, we split the image pairs based on the viewpoint change between them where $\theta_i^j$ means the error for image pairs whose relative viewpoint angle is between $i$ and $j$ degrees. 
One thing to note is that while two views with a relative angle of $180^\circ$ depict the opposite side of the object with no mutually visible surfaces, a room viewed from the opposite corner has a relative viewpoint change of $180^\circ$ with a large portion of the images being mutually visible. Hence, while increasing relative viewpoints imply increasing difficult, the numbers are not directly comparable, as one is viewing the scene from inside of it, but viewing the object from its outside.

\subsubsection{Semantic Correspondence}

\lsparagraph{Task Definition: } Given two images and a set of semantic keypoints in image one, the goal is to find the location of the pixel belonging to those key points in the second image. Key points are often semantic parts; \eg, a cat's left ear or the front right wheel of the car. Unlike the previous task where one has to find a set of points in both images that match each other, the set of points in the first image are already specified. Furthermore, while the previous task is matching points belonging to the same scene or the same object instance, semantic keypoints are defined at the class level so the images often depict two different instance; \eg, two different cats or two different cars. 

\lsparagraph{Inference Procedure: }
We follow previous work~\cite{tang2023dift,zhang2023tale} and simply use nearest neighbors. There is no need for filtering, as the goal is to just find the point in the second image that is most similar to the keypoint. 

\lsparagraph{Evaluation Metrics: }
The evaluation is often based on the percentage of keypoints within a pixel threshold; \ie, the percentage of predicted keypoints within $N$ pixels of the ground-truth match. 
The evaluation is based on the assumption that each keypoint has a single valid match in the second image. 
This results in each evaluation only considering the predicted keypoint and its ground-truth location, and ignoring everything else. 
We evaluate all models using this procedure in \cref{app:complete_results}, but we also consider an alternative evaluation as discussed in the main paper. 

An alternative way to evaluate the prediction is to compare to all keypoints in that image. Instead of asking how close the prediction is to the ground truth for the same keypoint type, we can ask which ground-truth keypoint is closest to the prediction. This allows us to understand which keypoints are getting confused with each other, rather than how many keypoints are correctly classified. This is important since the threshold is usually 10-20\% of the size of the bounding box, which can include several different keypoints. 

Previous work reports the average performance for all pairs. Instead, we separate the performance for image pairs of different viewpoint changes. Specifically, we use the annotation of viewpoint variation provided by SPair~\cite{min2019spair} and report the performance for different viewpoint difficulties.

%% file: supplemental/a_correlation.tex
One question tackled in our paper is how well performance is correlated across tasks. 
If several tasks are measuring the same capability, we would expect their performance to be well correlated.
Although a high correlation could be caused by a variety of other factors. 
Hence, while a high correlation provides some evidence that tasks measure the same capability, a very low correlation would imply that tasks are not related. 

We compute the correlation of model performance across different tasks and task domains. 
Specifically, we compute the Pearson correlation coefficient, which assumes a linear relationship between models. 
One possibility is that the relationships across tasks may not be linear and that a rank correlation might be well-suited. 
Empirically, we observe that both statistical measures often result in similar trends with some minor exceptions. 
When considering cases where the correlations deviate from each other, we find that they are caused by fluctuations in the rankings that arise for very small changes in performance for similarly performing models. 
As a result, we choose to report the Pearson correlation coefficient as a descriptive statistic of the relationships between model performance. 

When considering the overall model performance, such as Figure 1, we aggregate performance across all tasks. Since the absolute performance values are not directly comparable, we instead rely on the model rankings. While a model's ranking can fluctuate due to minor differences, such fluctuations tend to get canceled out when averaging the ranking across multiple tasks. We convert the rankings to a normalize rating where a 1 means the best performing model and 0 means the worst performance model. The overall ratings are shown in Figure 1. We emphasize that such ratings represent relative, not absolute, model performance. Hence, a rating of 1 does not mean the representations are 3D aware, but rather that they are more 3D aware than the other models considered.

%% file: supplemental/app_complete_results.tex
We chose to focus on general performance trends and salient comparisons in the main body of the paper. Here, we report the complete results for all the models and tasks considered in \Cref{tab:app_depth,tab:app_snorm,tab:corr_scannet,tab:corr_navi,tab:corr_spair}. 
Additionally, we attempt to provide some targeted comaprisons that shed some light on some of the findings discussed in the main paper, as well as address some of the confounders introduced by using publicly available checkpoints. 

\lsparagraph{What explains the low performance of vision and language models? }
One interesting finding is that vision and language models perform poorly across all tasks. 
Although previous work has shown that CLIP struggles with 2D spatial relationships~\cite{lewis2022does,li2024localizationvssemantics}, the disparity in performance that we observe is quite surprising. 
This makes it difficult to determine what causes the performance disparity. Below, we consider several possibilities: training data, training objective, model architecture, and augmentations.

One possibility is that CLIP's WIT dataset does not capture such relationships. However, we note that the OpenCLIP model trained on LAION and the SigLIP model trained on WebLI both achieve very similar performance as seen in \cref{tab:app_depth,tab:app_snorm,tab:corr_scannet,tab:corr_navi,tab:corr_spair}.
Furthermore, StableDiffusion achieves a very strong performance despite also being trained on LAION. 

Another possibility is that this is caused by the training objective; \ie, the contrastive objective discourages such relationships. 
Our experiments suggest that the training objective is likely the major culprit, but it remains unclear what about the objective is causing this issue. 
It is unclear whether semantics inhibits spatial understanding since both DeiT and StableDiffusion achieve a strong performance despite learning from some form of semantics. 
Furthermore, it is not the discriminative aspect of the learning objective, as DeiT and DINO are both trained with discriminative objects and also perform well. 
On the other hand, SigLIP learns from a non-contrastive objective and performs poorly. 

We find that the greatest improvement comes from changing the backbone from ViT to ConvNeXt.
This change results in a qualitative change in CLIP's performance; \eg, from predicting flat surfaces for depth to generating something that looks like a depth map. 
We note that this is not because ConvNeXt is a strictly superior architecture; \eg, we find that DeiT's ViT performs better than ConvNeXt for supervised training on ImageNet-22k. 

One final difference is the use of augmenations during training. Most of the approaches considered rely on augmentation to create a learning signal such as self-supervised learning or to provide additional diversity in the training data. However, CLIP relies on minimal augmentation, and we are unaware of any transformer-based checkpoints that train with augmentations. 
Instead, we consider two CLIP ConvNeXt checkpoints that were trained with and without AugReg~\cite{steiner2021augreg}. We find that augmentation boosts performance, but the gains are relatively small. It remains unclear if such augmentations would be more impactful when used with ViT backbones. 

In conclusion, we find that none of the factors individually explain the low performance of vision language models, although choice of architecture appears to be a useful direction to further puruse. Our findings also suggest that the training signal or model architecture does not independently determine the 3D awareness of features. However, more controlled experiments are needed to confirm this claim. 

\lsparagraph{What is the impact of model scale?}
While our experiments focused on ViT-B backbones, we also evaluated the large checkpoints and report their performance here. Although using a larger backbone typically increases performance, the improvement is fairly marginal. Furthermore, we find that the reported patterns and correlations do not change when we consider the larger checkpoints. 

\lsparagraph{What is the impact of model architecture?}
We choose a subset of our models and compare their performance when trained with ViT or ConvNeXt backbones. While ConvNeXts result in a huge performance gain for CLIP models, the results are mixed for other models; \eg, FCMAE and DeIT. As noted earlier, our experiments suggest that model architecture and training objective are not independent when it comes to 3D awareness. However, more research is needed to further explore this relationship. 

\lsparagraph{What is the impact of training dataset?}
While dataset scale appears to impact performance, the results are often mixed or marginal. Our main data point here is to compare models trained on ImageNet-1k and ImageNet-22k. Beyond scale, dataset curation appears to have a significant impact on performance as reported by \citet{oquab2023dinov2}. 
Despite these confounders, we find that the variation in performance due to different datasets is smaller than the variation caused by training objectives. 
Nevertheless, we hope that future work can shed some light on the impact of such datasets, and more importantly, on what properties of the dataset encourage or inhibit 3D awareness. 

\input{supplemental/fig_spair_additional}

\lsparagraph{Additional SPair Confusion Matrices. }
In the main body of the paper, we visualize the confusion matrices for the chair class under different viewpoint variations.
The chair class is most representative of this distinction for two reasons: (1) its keypoints neatly segment into semantic groups that only differ based on their relative location in the chair's canonical frame of references; and (2) semantically similar keypoints can be visible in the same image in different relative orientations for each other. 
Many other classes do not fulfill those criteria, especially the second point. 
For example, several keypoints for humans and animals are unique; \eg, mouth, nose, tail, forehead. 
In other cases, semantically related keypoints almost always appear in the same 2D configuration. 
For example, eyes and ears often appear in the same 2D configuration when they are both visible. As a result, their 2D relative locations are very strongly correlated with their identity, making it difficult to assess the 3D awareness of the model. 
This is confounded by common photographic biases in datasets; \eg, most human and naimal faces are often pictured facing the camera. 
Finally, many symmetric keypoints are almost never covisible. It is very rare for both front left and front right wheels to be covisible in the same image. 
Most car pictures featuring more than 1 visible wheel are side views. Hence, for most pairs of car views, a combination of coarse semantic class (\ie, car wheel) and relative location in the image can result in accurate correspondence. 

We observe these findings in the data. We present the confusion matrices StableDiffusion features on two additional classes: horses and airplanes. The results are shown in \cref{fig:spair_additional}. 
First, we find that unique classes, such as tail tip and tail base, have a similar accuracy across viewpoint variations. 
Furthermore, classes that appear in similar orientations, such as eyes or ears, are similarly unaffected. We note that the ear annotation refers to the front of the ear for horses.
Meanwhile, the top of the ear, which is also visible when the horse is looking away, exhibits a different behavior where it is far more confused when for larger viewpoint changes. 

We observe similar patterns for airplanes where unique keypoints such as the nose, windshield, and cockpit of the airplane are accurately predicted for large viewpoint changes. Meanwhile, the wheels, wings, and stabilizers are confused for larger viewpoint changes. 
To support that this is not simply due to some keypoints being more difficult, one can compare the performance for vertical stabilizers with the performance for right and left stabilizers. While the right and left stabilizers are confused, the vertical stabilizer, which looks the same but has a very different orientation, is still accurately predicted for larger viewpoint changes. 

Finally, we observe interesting trends for knees and hooves that appear to be confused regardless of viewpoint change. Since animals can deform, hooves and legs consistently appear in different orientations, especially for horses, which are often pictured while moving. Nevertheless, the confusion is strongly restricted to the semantically equivalent classes. This strongly suggests that while the model understands semantics, it lacks 3D awareness.

%% file: supplemental/fig_spair_additional.tex
\begin{figure*}
    \centering
    \includegraphics[width=\linewidth]{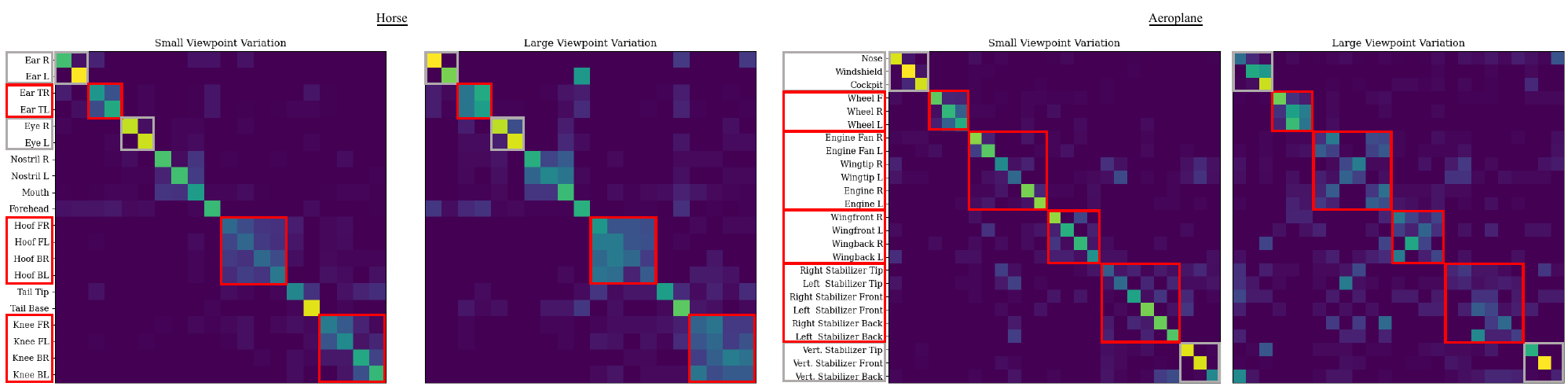}
    \caption{\textbf{Keypoint confusion matrices for horses and aeroplanes. }
    We find similar confusion patters in other SPair classes where large viewpoint changes result in high confusion between semantically related classes (highlighted in red).
    Furthermore, we find that keypoints that experience a lot of deformation (\eg, knees and hooves) are confused for all image pairs as they appear in different 2D locatons relative to each other. 
    However, we find that classes (highlighted in grey) that often appear in the same 2D configuration do not suffer from this effect. 
    }
    \label{fig:spair_additional}
\end{figure*}

%% file: supplemental/app_limitations.tex
Our goal is to understand to what extent current large-scale models ``understand'' the 3D world that images depict, as well as what factors encourage or discourage such understanding from emerging. 
This is very challenging, as our collective understanding of what models learn or how they represent what they learn is still very limited. 
Furthermore, there is no consensus on how 3D geometry should be represented, nor what it even means for a model to have 3D understanding. 
Finally, there are concrete challenges in evaluation that make it difficult to conduct controlled experiments. 
The study presented in this work is a first step toward answering these questions. 
While our experiments and analysis provide some initial answers to this question, our study suffers from several limitations that limit the strength of the conclusions we can draw and point to several avenues for future exploration. We discuss the limitations below and outline some open questions that we hope future work will address. 

\lsparagraph{Comparisons are limited to publicly available checkpoints. }
We focused on our analysis on publicly available checkpoints due to their availability and common use in the literature. 
However, as a result, our experiments often compare models trained with different recipes on different datasets. 
This is a significant confounding variable, as it is unclear if the trends we are observing are due to the main differentiators we observe or some minor implementation detail. 

Ideally, we would train the same backbone architecture on the same dataset with the same training recipe, but with different objectives as in \cite{elbanani2023lgssl,mu2021slip}. 
However, the computational resources needed to train and tune all models are prohibitively large. 
This problem is also likely to be exacerbated with the current trend of moving towards ever larger scales. 
Beyond the resources required, different approaches often have different data requirements such as curation, labels, captions, or dense annotations. 
There are currently no large-scale datasets that meet all those requirements. 

We tried to make comparisons more fair by choosing model checkpoints of comparable model capacity and pretraining data scale. 
Although we expected that the dataset or pre-training scale might end up dominating all other effects, our experiments suggest that other factors can be more important. 
For example, while CLIP is trained on a much larger dataset than DINO, DINO consistently outperforms CLIP. 
Furthermore, model performance does not seem to be very sensitive to training data, with CLIP achieving similar performance whether it was trained on WIT or LAION. 
We hope that our analysis identified interesting patterns and that future work can conduct much more controlled experiments that focus on a specific model comparison or dataset comparison. 

\lsparagraph{Our analysis focuses on two specific aspects of 3D awareness.} 
Our ability to perceive and infer 3D properties is remarkable. 
While many tasks can showcase this ability, we restrict our analysis to single-image surface reconstruction and multiview consistency. 
While those two aspects are fundamental to 3D understanding, they are not comprehensive. 
The ability to reconstruct the full 3D shape, predict deformation, and estimate physical properties such as support and containment fall under the general umbrella of 3D understanding. 
However, it is unclear which of those properties should be readily perceived from the image as opposed to inferred with more complex processing. 
While such delineations can be more philosophical in nature, they can guide the experimental design, as it is important to understand what we are looking for before designing experiments to find it. 
We expect that more comprehensive benchmarks of 3D understanding would measure such capabilities as well, and we hope that this work provides an initial step towards the study of this problem. 

\lsparagraph{Our experimental methodology focuses on probing methods.} 
Our analysis has focused on linear probe and zero-shot analysis approaches. We have done this to analyze the features as they are without changing them to better adapt to 3D tasks. While we argue that frozen features provide a more accurate understanding of the 3D awareness of the features, it would definitely be useful to understand how much of those patterns transfer to fine-tuning setups. Furthermore, if we consider recent advances in natural language processing, we see a rise in in-context learning with similar adaptations in computer vision. 
While linear probes could still be applied, it is likely that prompting-based methods will be more suited to analyze the 3D awareness of such models.

%% file: supplemental/tab_depth_all.tex
\begin{table*}[th!]
  \newcommand{\rowheader}{\rowcolor{Gray!20}}
  \centering
    \caption{
        \textbf{Depth Estimation Results. } We present the depth estimation results for all models. Models are grouped based on the supervisory signal.
        }
    \label{tab:app_depth}
  \setlength\tabcolsep{6pt}
  \footnotesize
  \begin{tabularx}{\linewidth}{Xll cccc cccc}
    \toprule
    & &
    & \multicolumn{4}{c}{NYU}  
    & \multicolumn{4}{c}{NAVI} 
    \\
    \cmidrule(lr){4-7} 
    \cmidrule(lr){8-11}
    \textbf{Model} & \textbf{Architecture} & \textbf{Dataset} &
    $\delta_1$ & $\delta_2$ & $\delta_3$ & RMSE &
    $\delta_1$ & $\delta_2$ & $\delta_3$ & RMSE \\
    \midrule
\midrule \rowheader \multicolumn{11}{l}{\textit{\textbf{Self-Supervised Models}}} \\
MAE~\cite{he2022masked}                 & ViT-B16   & IN-1k     &  67.89 &  91.43 &  97.92 & 0.6602    &  36.28 &  63.44 &  79.72 & 0.1568 \\
MAE~\cite{he2022masked}                 & ViT-L16   & IN-1k     &  70.22 &  92.50 &  98.00 & 0.6298    &  35.21 &  62.60 &  79.41 & 0.1588 \\
FCMAE~\cite{woo2023convnext}            & CNXTv2-B  & IN-22k    &  82.73 &  97.63 &  99.67 & 0.4860    &  48.61 &  76.15 &  88.44 & 0.1205 \\
DINO~\cite{caron2021emerging}           & ViT-B16   & IN-1k     &  80.44 &  96.52 &  99.26 & 0.5071    &  56.62 &  81.26 &  91.01 & 0.1043 \\
iBOT~\cite{zhou2021ibot}                & ViT-B16   & IN-1k     &  83.87 &  97.47 &  99.51 & 0.4635    &  57.11 &  82.00 &  91.49 & 0.1025 \\
iBOT~\cite{zhou2021ibot}                & ViT-B16   & IN-22k    &  85.04 &  97.82 &  99.55 & 0.4452    &  55.75 &  81.06 &  91.14 & 0.1044 \\
iBOT~\cite{zhou2021ibot}                & ViT-L16   & IN-1k     &  85.14 &  97.74 &  99.56 & 0.4451    &  62.56 &  85.27 &  93.19 & 0.0919 \\
iBOT~\cite{zhou2021ibot}                & ViT-L16   & IN-22k    &  91.08 &  98.98 &  99.79 & 0.3670    &  66.29 &  87.31 &  94.27 & 0.0833 \\
DINOv2~\cite{oquab2023dinov2}           & ViT-B14   & LVD       &  93.43 &  99.41 &  99.90 & 0.3307    &  68.84 &  89.13 &  95.31 & 0.0763 \\
DINOv2~\cite{oquab2023dinov2}           & ViT-L14   & LVD       &  94.84 &  99.57 &  99.93 & 0.3086    &  71.42 &  90.34 &  95.91 & 0.0721 \\
DINOv2 + reg~\cite{darcet2023vision}    & ViT-B14   & LVD       &  92.93 &  99.37 &  99.91 & 0.3355    &  66.56 &  87.94 &  94.74 & 0.0806 \\
\midrule \rowheader \multicolumn{11}{l}{\textit{\textbf{Classification-Supervised Models}}} \\
DeiT III~\cite{touvron2022deit}         & ViT-B16   & IN-22k    &  86.89 &  98.20 &  99.73 & 0.4240    &  59.74 &  83.78 &  92.61 & 0.0960 \\
DeiT III~\cite{touvron2022deit}         & ViT-L16   & IN-22k    &  89.55 &  98.75 &  99.78 & 0.3866    &  64.93 &  86.79 &  93.98 & 0.0855 \\
ConvNeXt~\cite{liu2022convnet}          & CNXT-B    & IN-22k    &  80.15 &  96.88 &  99.46 & 0.5105    &  43.63 &  71.60 &  85.49 & 0.1342 \\
\midrule \rowheader \multicolumn{11}{l}{\textit{\textbf{Vision Language Models}}} \\
CLIP\cite{radford2019language}          & ViT-B16   & WIT       &  52.11 &  81.70 &  93.72 & 0.9450    &  24.95 &  48.73 &  68.52 & 0.1993 \\
CLIP\cite{radford2019language}          & ViT-L14   & WIT       &  51.73 &  81.59 &  93.85 & 0.9445    &  23.97 &  46.71 &  66.34 & 0.2058 \\
CLIP\cite{openclip}                     & ViT-B16   & LAION     &  52.46 &  82.15 &  94.05 & 0.9351    &  24.33 &  47.43 &  67.31 & 0.2031 \\
CLIP\cite{openclip}                     & CNXT-B    & LAION     &  78.10 &  96.26 &  99.33 & 0.5285    &  45.24 &  73.35 &  86.68 & 0.1286 \\
CLIP + AugReg\cite{openclip}            & CNXT-B    & LAION     &  82.14 &  97.24 &  99.48 & 0.4784    &  48.22 &  75.85 &  88.09 & 0.1211 \\
SigLIP~\cite{zhai2023sigmoid}           & ViT-B16   & WebLI     &  63.81 &  89.72 &  97.33 & 0.7187    &  36.49 &  63.07 &  79.20 & 0.1571 \\
SigLIP~\cite{zhai2023sigmoid}           & ViT-L16   & WebLI     &  65.26 &  90.53 &  97.78 & 0.6931    &  35.94 &  62.43 &  78.78 & 0.1588 \\
\midrule \rowheader \multicolumn{11}{l}{\textit{\textbf{Text-Conditioned Image Generation Models}}} \\
StableDiffusion~\cite{rombach2022high}  & UNet      & LAION     &  82.60 &  97.05 &  99.48 & 0.4801    &  51.86 &  78.68 &  89.77 & 0.1094 \\
\midrule \rowheader \multicolumn{11}{l}{\textit{\textbf{Densely-Supervised Models}}} \\
SAM~\cite{kirillov2023sam}              & ViT-B16   & SA-1B     &  75.60 &  95.00 &  98.94 & 0.5665    &  49.28 &  76.36 &  88.40 & 0.1206 \\
SAM~\cite{kirillov2023sam}              & ViT-L16   & SA-1B     &  81.57 &  97.12 &  99.49 & 0.4905    &  52.45 &  79.17 &  90.25 & 0.1105 \\
MiDaS~\cite{ranftl2021dpt}              & ViT-L16   & MIX 6     &  78.65 &  96.05 &  98.99 & 0.5300    &  58.53 &  82.58 &  91.60 & 0.1001 \\

\bottomrule
\end{tabularx}
\end{table*}

%% file: supplemental/tab_snorm_all.tex
\begin{table*}[!ht]
  \newcommand{\rowheader}{\rowcolor{Gray!20}}
  \centering
    \caption{
  \textbf{Surface Normal Estimation Results. } We present the surface normal estimation results for all models. Models are grouped based on the supervisory signal.}
  \label{tab:app_snorm}
  \setlength\tabcolsep{6pt}
  \footnotesize
  \begin{tabularx}{\linewidth}{Xll cccc cccc}
    \toprule
    & &
    & \multicolumn{4}{c}{NYU}  
    & \multicolumn{4}{c}{NAVI} 
    \\
    \cmidrule(lr){4-7} 
    \cmidrule(lr){8-11}    
    \textbf{Model} & \textbf{Architecture} & \textbf{Dataset} &
    $11.25^\circ$ & $22.5^\circ$ & $30^\circ$ & RMSE &
    $11.25^\circ$ & $22.5^\circ$ & $30^\circ$ & RMSE \\
    \midrule
\midrule \rowheader \multicolumn{11}{l}{\textit{\textbf{Self-Supervised Models}}} \\
MAE~\cite{he2022masked}                 & ViT-B16   & IN-1k     &  44.60 &  66.08 &  74.37 &  30.34    &  30.27 &  57.70 &  69.97 &  32.12 \\
MAE~\cite{he2022masked}                 & ViT-L16   & IN-1k     &  45.60 &  67.50 &  75.67 &  29.57    &  30.36 &  58.20 &  70.54 &  31.59 \\
FCMAE~\cite{woo2023convnext}            & CNXTv2-B  & IN-22k    &  43.87 &  68.73 &  78.05 &  27.24    &  28.91 &  58.06 &  70.76 &  31.54 \\
DINO~\cite{caron2021emerging}           & ViT-B16   & IN-1k     &  49.11 &  69.32 &  77.02 &  28.35    &  39.56 &  65.85 &  76.23 &  28.75 \\
iBOT~\cite{zhou2021ibot}                & ViT-B16   & IN-1k     &  52.60 &  72.52 &  79.60 &  26.89    &  40.65 &  66.79 &  76.99 &  28.32 \\
iBOT~\cite{zhou2021ibot}                & ViT-B16   & IN-22k    &  54.54 &  73.79 &  80.52 &  26.10    &  22.75 &  50.60 &  64.10 &  35.03 \\
iBOT~\cite{zhou2021ibot}                & ViT-L16   & IN-1k     &  54.53 &  74.30 &  81.06 &  25.78    &  43.45 &  69.20 &  78.84 &  27.24 \\
iBOT~\cite{zhou2021ibot}                & ViT-L16   & IN-22k    &  58.51 &  76.68 &  82.90 &  24.46    &  45.19 &  70.51 &  79.84 &  26.56 \\
DINOv2~\cite{oquab2023dinov2}           & ViT-B14   & LVD       &  62.01 &  79.32 &  85.32 &  22.41    &  47.55 &  72.92 &  81.89 &  25.27 \\
DINOv2~\cite{oquab2023dinov2}           & ViT-L14   & LVD       &  64.05 &  80.78 &  86.48 &  21.55    &  50.15 &  74.70 &  83.12 &  24.52 \\
DINOv2 + reg~\cite{darcet2023vision}    & ViT-B14   & LVD       &  61.37 &  79.12 &  85.22 &  22.54    &  45.81 &  72.00 &  81.28 &  25.66 \\
\midrule \rowheader \multicolumn{11}{l}{\textit{\textbf{Classification-Supervised Models}}} \\
DeiT III~\cite{touvron2022deit}         & ViT-B16   & IN-22k    &  39.58 &  64.22 &  74.07 &  29.44    &  26.91 &  55.66 &  68.71 &  32.47 \\
DeiT III~\cite{touvron2022deit}         & ViT-L16   & IN-22k    &  55.73 &  75.34 &  82.00 &  25.05    &  32.91 &  61.54 &  73.49 &  29.92 \\
ConvNeXt~\cite{liu2022convnet}          & CNXT-B    & IN-22k    &  25.79 &  47.42 &  59.15 &  35.40    &  21.72 &  50.54 &  64.75 &  34.34 \\
\midrule \rowheader \multicolumn{11}{l}{\textit{\textbf{Vision Language Models}}} \\
CLIP\cite{radford2019language}          & ViT-B16   & WIT       &  28.85 &  51.93 &  63.07 &  35.34    &  17.01 &  42.06 &  56.10 &  39.58 \\
CLIP\cite{radford2019language}          & ViT-L14   & WIT       &  28.07 &  50.74 &  61.97 &  35.84    &  15.03 &  38.87 &  52.93 &  41.41 \\
CLIP\cite{openclip}                     & ViT-B16   & LAION     &  30.11 &  53.65 &  64.60 &  34.68    &  16.23 &  40.95 &  55.07 &  39.90 \\
CLIP\cite{openclip}                     & CNXT-B    & LAION     &  43.54 &  67.93 &  77.06 &  27.88    &  31.28 &  59.41 &  71.60 &  31.08 \\
CLIP + AugReg\cite{openclip}            & CNXT-B    & LAION     &  45.50 &  69.53 &  78.43 &  27.07    &  34.17 &  62.29 &  73.87 &  29.82 \\
SigLIP~\cite{zhai2023sigmoid}           & ViT-B16   & WebLI     &  30.68 &  52.73 &  63.60 &  34.96    &  21.47 &  47.60 &  60.93 &  36.69 \\
SigLIP~\cite{zhai2023sigmoid}           & ViT-L16   & WebLI     &  31.68 &  54.11 &  64.97 &  34.20    &  21.24 &  46.83 &  60.19 &  37.01 \\
\midrule \rowheader \multicolumn{11}{l}{\textit{\textbf{Text-Conditioned Image Generation Models}}} \\
StableDiffusion~\cite{rombach2022high}  & UNet      & LAION     &  58.29 &  76.28 &  82.64 &  24.68    &  40.31 &  67.18 &  77.55 &  27.86 \\
\midrule \rowheader \multicolumn{11}{l}{\textit{\textbf{Densely-Supervised Models}}} \\
SAM~\cite{kirillov2023sam}              & ViT-B16   & SA-1B     &  46.37 &  70.40 &  78.98 &  26.89    &  36.66 &  64.25 &  75.36 &  29.01 \\
SAM~\cite{kirillov2023sam}              & ViT-L16   & SA-1B     &  52.03 &  74.61 &  82.20 &  24.87    &  39.23 &  66.87 &  77.59 &  27.66 \\
MiDaS~\cite{ranftl2021dpt}              & ViT-L16   & MIX 6     &  49.58 &  69.82 &  77.17 &  28.51    &  39.94 &  66.36 &  76.66 &  28.54 \\

\bottomrule
\end{tabularx}
\end{table*}

%% file: supplemental/tab_corr_scannet.tex
\begin{table*}
  \newcommand{\rowheader}{\rowcolor{Gray!20}}
  \newcommand{\hlc}{\cellcolor{Orange!20}}

  \centering    
  \caption{\textbf{Correspondence Estimation Results for ScanNet. } We present the ScanNet correspondence estimation results for all models. The results are presented for features extracted at different layers with performance binned for different relative viewpoint changes between image pairs. The highest performing set of results for each model are highlighted and bolded. }
  \label{tab:corr_scannet}
  \setlength\tabcolsep{3pt}
  \scriptsize
  \begin{tabularx}{\linewidth}{Xll@{\hskip 10pt} rrrr@{\hskip 10pt} rrrr@{\hskip 10pt} rrrr@{\hskip 10pt} rrrr}
    \toprule
    & &
    & \multicolumn{4}{c}{Block$_0$}
    & \multicolumn{4}{c}{Block$_1$}
    & \multicolumn{4}{c}{Block$_2$}
    & \multicolumn{4}{c}{Block$_3$}
    \\
    \cmidrule(lr){4-7} 
    \cmidrule(lr){8-11}
    \cmidrule(lr){12-15}
    \cmidrule(lr){16-19}    
    \textbf{Model} & \textbf{Architecture} & \textbf{Dataset} &
    $\theta_{0}^{15}$ & $\theta_{15}^{30}$ & $\theta_{30}^{60}$ & $\theta_{60}^{180}$ &
    $\theta_{0}^{15}$ & $\theta_{15}^{30}$ & $\theta_{30}^{60}$ & $\theta_{60}^{180}$ &
    $\theta_{0}^{15}$ & $\theta_{15}^{30}$ & $\theta_{30}^{60}$ & $\theta_{60}^{180}$ &
    $\theta_{0}^{15}$ & $\theta_{15}^{30}$ & $\theta_{30}^{60}$ & $\theta_{60}^{180}$\\
    \midrule
\midrule \rowheader \multicolumn{19}{l}{\textit{\textbf{Self-Supervised Models}}} \\
MAE~\cite{he2022masked}                 & ViT-B16   & IN-1k       &  3.4 &  2.8 &  3.8 &  2.3   &  4.7 &  3.6 &  3.9 &  2.6   &  7.8 &  5.5 &  4.9 &  3.0   & 12.5 &  8.0 &  6.0 &  3.6  \\
MAE~\cite{he2022masked}                 & ViT-L16   & IN-1k       &  5.2 &  3.8 &  4.1 &  2.6   &  8.2 &  6.3 &  5.2 &  3.1   & 14.1 & 10.1 &  7.0 &  4.1   & 15.6 & 10.9 &  7.5 &  4.3  \\
FCMAE~\cite{woo2023convnext}            & CNXTv2-B  & IN-22k      & 26.2 & 19.8 & 12.1 &  5.5   & 60.8 & 49.7 & 28.1 &  9.8   & 31.6 & 22.4 & 13.2 &  5.8   & 36.8 & 26.2 & 14.9 &  7.2  \\
DINO~\cite{caron2021emerging}           & ViT-B16   & IN-1k       & 14.3 &  9.5 &  7.9 &  3.9   & 42.8 & 32.4 & 21.0 &  9.1   & 44.5 & 34.1 & 22.2 &  9.8   & 45.0 & 34.3 & 22.6 & 10.7  \\
iBOT~\cite{zhou2021ibot}                & ViT-B16   & IN-1k       & 19.3 & 13.0 &  9.5 &  4.1   & 37.2 & 26.6 & 17.6 &  7.8   & 39.0 & 28.7 & 18.9 &  8.8   & 37.8 & 27.4 & 18.4 &  9.1  \\
iBOT~\cite{zhou2021ibot}                & ViT-B16   & IN-22k      & 14.2 & 10.1 &  7.8 &  3.6   & 27.9 & 19.2 & 12.6 &  6.2   & 36.0 & 25.8 & 16.5 &  8.0   & 30.1 & 20.7 & 13.9 &  7.0  \\
iBOT~\cite{zhou2021ibot}                & ViT-L16   & IN-1k       & 30.3 & 20.7 & 14.0 &  5.8   & 43.8 & 32.8 & 21.0 &  9.1   & 44.7 & 33.8 & 22.0 & 10.1   & 46.0 & 34.6 & 22.8 & 10.8  \\
iBOT~\cite{zhou2021ibot}                & ViT-L16   & IN-22k      & 27.9 & 20.0 & 13.1 &  5.5   & 42.2 & 31.1 & 19.9 &  8.8   & 41.3 & 30.5 & 19.6 &  9.3   & 40.4 & 29.9 & 20.3 & 10.2  \\
DINOv2~\cite{oquab2023dinov2}           & ViT-B14   & LVD         & 25.4 & 19.5 & 12.7 &  4.9   & 47.1 & 36.4 & 22.4 &  8.4   & 37.6 & 26.8 & 16.8 &  7.5   & 37.0 & 27.5 & 19.7 & 11.2  \\
DINOv2~\cite{oquab2023dinov2}           & ViT-L14   & LVD         & 12.6 & 10.4 &  7.8 &  4.0   & 27.6 & 19.4 & 12.5 &  4.7   & 34.7 & 23.7 & 15.8 &  6.4   & 36.4 & 26.8 & 20.4 & 12.1  \\
DINOv2 + reg~\cite{darcet2023vision}    & ViT-B14   & LVD         & 29.0 & 24.6 & 15.1 &  5.8   & 56.1 & 47.3 & 29.5 & 10.3   & 48.4 & 37.8 & 24.2 & 10.0   & 41.9 & 33.6 & 23.2 & 12.2  \\
\midrule \rowheader \multicolumn{19}{l}{\textit{\textbf{Classification-Supervised Models}}} \\
DeiT III~\cite{touvron2022deit}         & ViT-B16   & IN-22k      & 17.6 & 12.1 &  8.8 &  3.4   & 38.3 & 29.7 & 17.9 &  7.2   & 28.5 & 21.3 & 14.0 &  6.6   & 20.7 & 13.7 &  9.2 &  5.0  \\
DeiT III~\cite{touvron2022deit}         & ViT-L16   & IN-22k      & 34.3 & 26.8 & 16.6 &  5.4   & 36.4 & 27.8 & 16.7 &  7.0   & 26.5 & 19.4 & 12.7 &  6.3   & 28.3 & 20.6 & 14.0 &  7.5  \\
ConvNeXt~\cite{liu2022convnet}          & CNXT-B    & IN-22k      & 22.8 & 17.1 &  9.9 &  4.7   & 42.9 & 32.9 & 18.1 &  6.9   & 10.2 &  6.7 &  3.9 &  2.6   & 15.0 & 10.6 &  5.7 &  3.6  \\
\midrule \rowheader \multicolumn{19}{l}{\textit{\textbf{Vision Language Models}}} \\
CLIP\cite{radford2019language}          & ViT-B16   & WIT         & 10.5 &  8.3 &  6.5 &  4.4   &  3.7 &  3.3 &  3.8 &  2.6   &  3.0 &  2.4 &  3.1 &  2.1   &  2.5 &  2.1 &  2.7 &  1.8  \\
CLIP\cite{radford2019language}          & ViT-L14   & WIT         &  8.4 &  7.3 &  5.8 &  3.9   &  4.8 &  4.1 &  4.3 &  2.9   &  3.9 &  3.3 &  3.6 &  2.6   &  3.4 &  2.9 &  3.0 &  2.5  \\
CLIP\cite{openclip}                     & ViT-B16   & LAION       & 15.5 & 11.6 &  8.3 &  4.6   &  5.8 &  4.8 &  4.7 &  3.2   &  2.5 &  2.3 &  3.0 &  2.1   &  2.5 &  2.1 &  2.7 &  2.0  \\
CLIP\cite{openclip}                     & CNXT-B    & LAION       & 26.8 & 20.8 & 13.3 &  6.6   & 30.3 & 21.7 & 12.1 &  5.3   & 47.1 & 38.1 & 22.3 &  8.9   & 37.1 & 28.5 & 16.6 &  8.0  \\
CLIP + AugReg\cite{openclip}            & CNXT-B    & LAION       & 24.2 & 18.9 & 12.4 &  6.3   & 37.7 & 27.3 & 14.5 &  6.0   & 32.3 & 22.9 & 13.0 &  5.4   & 31.6 & 22.8 & 13.1 &  7.1  \\
SigLIP~\cite{zhai2023sigmoid}           & ViT-B16   & WebLI       & 17.0 & 12.5 &  9.2 &  5.9   & 16.6 & 12.0 &  8.9 &  5.5   & 15.8 & 11.3 &  8.6 &  5.4   & 14.2 & 10.4 &  8.1 &  5.3  \\
SigLIP~\cite{zhai2023sigmoid}           & ViT-L16   & WebLI       & 15.4 & 11.1 &  8.1 &  4.9   & 14.4 & 10.2 &  7.3 &  4.6   & 13.4 &  9.3 &  6.7 &  4.2   & 11.7 &  8.5 &  6.4 &  4.2  \\
\midrule \rowheader \multicolumn{19}{l}{\textit{\textbf{Text-Conditioned Image Generation Models}}} \\
StableDiffusion~\cite{rombach2022high}  & UNet      & LAION       & 10.2 &  5.0 &  3.0 &  1.4   & 66.4 & 55.0 & 31.0 &  8.4   & 60.4 & 48.3 & 27.4 &  9.1   & 14.4 & 10.9 &  7.7 &  4.5  \\
\midrule \rowheader \multicolumn{19}{l}{\textit{\textbf{Densely-Supervised Models}}} \\
SAM~\cite{kirillov2023sam}              & ViT-B16   & SA-1B       &  8.3 &  6.0 &  5.6 &  2.9   & 35.0 & 26.2 & 17.4 &  5.3   & 52.9 & 44.7 & 29.3 &  8.9   & 55.3 & 47.0 & 30.4 &  9.4  \\
SAM~\cite{kirillov2023sam}              & ViT-L16   & SA-1B       & 14.5 &  9.9 &  7.5 &  3.5   & 37.2 & 29.7 & 19.7 &  6.2   & 47.6 & 40.4 & 27.3 &  8.7   & 52.6 & 43.9 & 28.7 &  9.6  \\
MiDaS~\cite{ranftl2021dpt}              & ViT-L16   & MIX 6       & 50.3 & 39.0 & 24.4 & 11.2   & 56.4 & 47.4 & 31.6 & 13.9   & 55.5 & 46.0 & 30.8 & 14.3   & 52.4 & 42.1 & 27.6 & 13.1  \\

\bottomrule
\end{tabularx}
\end{table*}

%% file: supplemental/tab_corr_navi.tex
\begin{table*}
  \newcommand{\rowheader}{\rowcolor{Gray!20}}
  \newcommand{\hlc}{\cellcolor{Orange!20}}

  \centering
    \caption{\textbf{Correspondence Estimation Results for NAVI. } We present the NAVI correspondence estimation results for all models. The results are presented for features extracted at different layers with performance binned for different relative viewpoint changes between image pairs. The highest performing set of results for each model are highlighted and bolded. }
    
  \label{tab:corr_navi}
  \setlength\tabcolsep{3pt}
  \scriptsize
  \begin{tabularx}{\linewidth}{Xll@{\hskip 10pt} rrrr@{\hskip 10pt} rrrr@{\hskip 10pt} rrrr@{\hskip 10pt} rrrr}
    \toprule
    & &
    & \multicolumn{4}{c}{Block$_0$}
    & \multicolumn{4}{c}{Block$_1$}
    & \multicolumn{4}{c}{Block$_2$}
    & \multicolumn{4}{c}{Block$_3$}
    \\
    \cmidrule(lr){4-7} 
    \cmidrule(lr){8-11}
    \cmidrule(lr){12-15}
    \cmidrule(lr){16-19}    
    \textbf{Model} & \textbf{Architecture} & \textbf{Dataset} &
    $\theta_{0}^{30}$ & $\theta_{30}^{60}$ & $\theta_{60}^{90}$ & $\theta_{90}^{120}$ &
    $\theta_{0}^{30}$ & $\theta_{30}^{60}$ & $\theta_{60}^{90}$ & $\theta_{90}^{120}$ &
    $\theta_{0}^{30}$ & $\theta_{30}^{60}$ & $\theta_{60}^{90}$ & $\theta_{90}^{120}$ &
    $\theta_{0}^{30}$ & $\theta_{30}^{60}$ & $\theta_{60}^{90}$ & $\theta_{90}^{120}$\\
    \midrule

\midrule \rowheader \multicolumn{19}{l}{\textit{\textbf{Self-Supervised Models}}} \\
MAE~\cite{he2022masked}                 & ViT-B16   & IN-1k       & 73.1 & 40.9 & 19.2 & 10.7   & 75.8 & 42.3 & 19.8 & 10.8   & 76.9 & 43.1 & 20.3 & 11.1   & 76.3 & 43.0 & 20.2 & 11.1  \\
MAE~\cite{he2022masked}                 & ViT-L16   & IN-1k       & 68.5 & 38.7 & 19.3 & 11.2   & 73.2 & 40.7 & 19.8 & 11.0   & 74.0 & 40.9 & 19.8 & 10.9   & 73.3 & 40.3 & 19.4 & 10.7  \\
FCMAE~\cite{woo2023convnext}            & CNXTv2-B  & IN-22k      & 37.9 & 24.7 & 16.3 & 11.0   & 66.2 & 37.5 & 19.5 & 11.6   & 74.0 & 48.4 & 28.6 & 18.1   & 82.8 & 59.5 & 39.6 & 26.7  \\
DINO~\cite{caron2021emerging}           & ViT-B16   & IN-1k       & 85.7 & 50.2 & 22.7 & 11.9   & 89.3 & 58.1 & 30.1 & 18.3   & 87.2 & 57.4 & 31.7 & 20.6   & 86.0 & 56.0 & 31.3 & 20.3  \\
iBOT~\cite{zhou2021ibot}                & ViT-B16   & IN-1k       & 85.3 & 49.3 & 22.2 & 11.7   & 90.3 & 57.1 & 27.9 & 16.1   & 89.4 & 59.1 & 31.6 & 20.2   & 89.0 & 58.3 & 32.5 & 22.2  \\
iBOT~\cite{zhou2021ibot}                & ViT-B16   & IN-22k      & 84.3 & 47.6 & 21.5 & 11.3   & 90.3 & 55.4 & 25.7 & 15.1   & 89.9 & 59.6 & 31.8 & 20.0   & 88.7 & 57.7 & 31.2 & 20.8  \\
iBOT~\cite{zhou2021ibot}                & ViT-L16   & IN-1k       & 90.2 & 56.2 & 26.3 & 14.4   & 92.2 & 63.0 & 32.7 & 20.4   & 91.4 & 63.3 & 34.6 & 22.8   & 90.3 & 63.6 & 36.7 & 25.4  \\
iBOT~\cite{zhou2021ibot}                & ViT-L16   & IN-22k      & 89.3 & 54.1 & 24.1 & 12.6   & 93.1 & 65.5 & 34.0 & 20.9   & 92.2 & 66.2 & 36.7 & 23.9   & 89.5 & 64.5 & 39.0 & 27.0  \\
DINOv2~\cite{oquab2023dinov2}           & ViT-B14   & LVD         & 79.4 & 46.1 & 22.2 & 12.3   & 93.6 & 63.7 & 29.4 & 14.9   & 94.4 & 68.8 & 36.8 & 20.5   & 90.6 & 69.3 & 45.9 & 32.0  \\
DINOv2~\cite{oquab2023dinov2}           & ViT-L14   & LVD         & 66.2 & 37.2 & 19.6 & 11.5   & 92.1 & 57.9 & 25.6 & 12.8   & 95.3 & 70.0 & 35.4 & 18.5   & 92.2 & 72.3 & 48.9 & 35.0  \\
DINOv2 + reg~\cite{darcet2023vision}    & ViT-B14   & LVD         & 73.0 & 41.9 & 21.0 & 12.0   & 92.6 & 62.0 & 28.7 & 14.6   & 94.4 & 70.0 & 37.9 & 21.3   & 89.0 & 67.3 & 44.8 & 31.1  \\
\midrule \rowheader \multicolumn{19}{l}{\textit{\textbf{Classification-Supervised Models}}} \\
DeiT III~\cite{touvron2022deit}         & ViT-B16   & IN-22k      & 88.6 & 51.8 & 22.9 & 12.1   & 91.5 & 62.8 & 34.3 & 22.0   & 84.3 & 58.0 & 37.9 & 26.7   & 62.7 & 38.5 & 24.6 & 16.7  \\
DeiT III~\cite{touvron2022deit}         & ViT-L16   & IN-22k      & 88.7 & 54.2 & 24.5 & 13.3   & 92.3 & 65.5 & 36.9 & 24.1   & 86.1 & 60.6 & 39.2 & 27.4   & 76.8 & 50.8 & 32.9 & 22.5  \\
ConvNeXt~\cite{liu2022convnet}          & CNXT-B    & IN-22k      & 39.1 & 24.3 & 15.6 & 10.2   & 49.2 & 25.8 & 14.8 &  9.4   & 75.8 & 46.1 & 26.6 & 16.6   & 80.3 & 52.2 & 31.5 & 20.3  \\
\midrule \rowheader \multicolumn{19}{l}{\textit{\textbf{Vision Language Models}}} \\
CLIP\cite{radford2019language}          & ViT-B16   & WIT         & 42.3 & 26.2 & 16.2 & 10.7   & 34.8 & 22.8 & 14.6 & 10.0   & 25.7 & 17.9 & 12.9 &  8.9   & 22.1 & 15.6 & 11.4 &  7.9  \\
CLIP\cite{radford2019language}          & ViT-L14   & WIT         & 36.3 & 22.6 & 14.8 & 10.1   & 28.1 & 18.9 & 13.6 &  9.4   & 21.3 & 15.7 & 12.1 &  8.6   & 17.9 & 13.6 & 10.9 &  7.7  \\
CLIP\cite{openclip}                     & ViT-B16   & LAION       & 41.8 & 25.2 & 15.8 & 10.5   & 36.5 & 22.6 & 14.5 &  9.7   & 26.0 & 17.1 & 12.2 &  8.6   & 21.6 & 15.3 & 11.0 &  7.9  \\
CLIP\cite{openclip}                     & CNXT-B    & LAION       & 34.1 & 22.9 & 15.5 & 10.9   & 45.0 & 27.6 & 17.1 & 11.1   & 84.7 & 56.9 & 32.2 & 19.6   & 76.7 & 47.1 & 30.0 & 20.5  \\
CLIP + AugReg\cite{openclip}            & CNXT-B    & LAION       & 34.4 & 23.0 & 15.1 & 10.4   & 48.1 & 28.0 & 16.7 & 10.6   & 84.5 & 54.4 & 30.7 & 18.1   & 78.9 & 50.1 & 31.6 & 21.4  \\
SigLIP~\cite{zhai2023sigmoid}           & ViT-B16   & WebLI       & 40.6 & 26.7 & 18.6 & 12.7   & 41.7 & 27.5 & 18.6 & 12.6   & 40.5 & 26.5 & 18.3 & 12.3   & 35.3 & 23.2 & 16.7 & 11.4  \\
SigLIP~\cite{zhai2023sigmoid}           & ViT-L16   & WebLI       & 39.0 & 25.8 & 17.5 & 11.8   & 39.7 & 26.1 & 17.6 & 11.9   & 36.6 & 24.2 & 16.6 & 11.4   & 29.9 & 19.8 & 14.9 & 10.2  \\
\midrule \rowheader \multicolumn{19}{l}{\textit{\textbf{Text-Conditioned Image Generation Models}}} \\
StableDiffusion~\cite{rombach2022high}  & UNet      & LAION       & 77.4 & 36.4 & 14.8 &  7.4   & 91.2 & 58.9 & 25.7 & 11.1   & 70.8 & 41.1 & 20.3 & 11.5   & 33.9 & 21.9 & 14.7 &  9.7  \\
\midrule \rowheader \multicolumn{19}{l}{\textit{\textbf{Densely-Supervised Models}}} \\
SAM~\cite{kirillov2023sam}              & ViT-B16   & SA-1B       & 77.8 & 42.7 & 19.9 & 11.4   & 83.0 & 48.4 & 22.0 & 11.9   & 88.6 & 56.2 & 25.0 & 12.9   & 88.2 & 56.5 & 25.3 & 12.7  \\
SAM~\cite{kirillov2023sam}              & ViT-L16   & SA-1B       & 78.0 & 43.3 & 20.4 & 11.4   & 86.4 & 52.0 & 23.8 & 12.5   & 91.2 & 60.1 & 28.2 & 14.2   & 88.5 & 57.6 & 26.9 & 13.5  \\
MiDaS~\cite{ranftl2021dpt}              & ViT-L16   & MIX 6       & 79.0 & 49.1 & 25.0 & 14.5   & 83.2 & 56.0 & 32.1 & 21.6   & 82.2 & 56.3 & 33.1 & 22.9   & 79.6 & 53.0 & 31.4 & 21.6  \\

\bottomrule
\end{tabularx}
\end{table*}

%% file: supplemental/tab_corr_spair.tex
\begin{table*}
  \newcommand{\rowheader}{\rowcolor{Gray!20}}
  \newcommand{\hlc}{\cellcolor{Orange!20}}
  \centering

  \caption{\textbf{Correspondence Estimation Results for SPair-71k. } We present the  SPair-71k correspondence estimation results for all models. The results are presented for features extracted at different layers with performance binned based on the viewpoint variation for the image pair. The highest performing set of results for each model are highlighted and bolded. }  
  \label{tab:corr_spair}
  \setlength\tabcolsep{3pt}
  \scriptsize
  \begin{tabularx}{\linewidth}{Xll@{\hskip 12pt} rrrr@{\hskip 12pt} rrrr@{\hskip 12pt} rrrr@{\hskip 12pt} rrrr}
    \toprule
    & &
    & \multicolumn{4}{c}{Block$_0$}
    & \multicolumn{4}{c}{Block$_1$}
    & \multicolumn{4}{c}{Block$_2$}
    & \multicolumn{4}{c}{Block$_3$}
    \\
    \cmidrule(lr){4-7} 
    \cmidrule(lr){8-11}
    \cmidrule(lr){12-15}
    \cmidrule(lr){16-19}    
    \textbf{Model} & \textbf{Architecture} & \textbf{Dataset} &
    $d{=}0$ & $d{=}1$ & $d{=}2$ & all &
    $d{=}0$ & $d{=}1$ & $d{=}2$ & all &
    $d{=}0$ & $d{=}1$ & $d{=}2$ & all &
    $d{=}0$ & $d{=}1$ & $d{=}2$ & all \\
    \midrule
\midrule \rowheader \multicolumn{19}{l}{\textit{\textbf{Self-Supervised Models}}} \\
MAE~\cite{he2022masked}                 & ViT-B16   & IN-1k       &  8.3 &  4.7 &  3.8 &  6.8   &  9.9 &  6.3 &  5.1 &  7.9   & 10.5 &  6.6 &  5.5 &  8.5   &  9.8 &  6.2 &  4.9 &  7.9  \\
MAE~\cite{he2022masked}                 & ViT-L16   & IN-1k       &  8.3 &  5.5 &  4.6 &  6.9   &  9.1 &  6.5 &  5.1 &  7.6   &  9.3 &  6.4 &  5.2 &  7.7   &  8.3 &  5.8 &  4.5 &  7.2  \\
FCMAE~\cite{woo2023convnext}            & CNXTv2-B  & IN-22k      &  5.3 &  4.7 &  4.9 &  5.0   &  8.3 &  6.7 &  6.0 &  7.3   & 26.4 & 24.3 & 25.4 & 24.8   & 28.9 & 27.1 & 28.2 & 27.5  \\
DINO~\cite{caron2021emerging}           & ViT-B16   & IN-1k       & 15.3 &  8.2 &  6.4 & 11.6   & 26.7 & 17.9 & 18.6 & 21.9   & 32.1 & 25.1 & 25.9 & 28.3   & 30.4 & 24.0 & 24.3 & 26.8  \\
iBOT~\cite{zhou2021ibot}                & ViT-B16   & IN-1k       & 14.3 &  7.3 &  5.5 & 10.4   & 25.3 & 14.6 & 14.0 & 20.0   & 35.5 & 25.3 & 25.5 & 30.6   & 39.9 & 30.3 & 32.3 & 35.7  \\
iBOT~\cite{zhou2021ibot}                & ViT-B16   & IN-22k      & 12.7 &  6.6 &  4.5 &  9.6   & 22.0 & 12.5 & 10.6 & 17.0   & 36.0 & 25.0 & 24.6 & 30.6   & 41.9 & 29.6 & 31.1 & 36.1  \\
iBOT~\cite{zhou2021ibot}                & ViT-L16   & IN-1k       & 22.4 & 12.0 & 10.0 & 17.1   & 39.3 & 25.6 & 26.1 & 32.8   & 44.9 & 32.8 & 33.7 & 39.3   & 48.9 & 36.2 & 38.5 & 43.2  \\
iBOT~\cite{zhou2021ibot}                & ViT-L16   & IN-22k      & 19.4 &  9.9 &  7.8 & 14.7   & 39.6 & 23.2 & 23.5 & 31.6   & 46.8 & 32.2 & 33.6 & 40.0   & 51.2 & 40.9 & 43.1 & 45.9  \\
DINOv2~\cite{oquab2023dinov2}           & ViT-B14   & LVD         & 13.0 &  8.4 &  7.1 & 10.3   & 35.9 & 20.9 & 16.3 & 28.0   & 55.8 & 37.0 & 34.4 & 46.7   & 62.4 & 51.9 & 53.3 & 56.8  \\
DINOv2~\cite{oquab2023dinov2}           & ViT-L14   & LVD         &  8.5 &  6.2 &  5.3 &  7.5   & 25.0 & 14.0 & 10.8 & 19.3   & 53.9 & 34.6 & 31.6 & 44.5   & 62.8 & 53.3 & 54.2 & 57.2  \\
DINOv2 + reg~\cite{darcet2023vision}    & ViT-B14   & LVD         & 12.0 &  8.3 &  7.3 & 10.1   & 33.2 & 19.6 & 15.4 & 26.1   & 57.4 & 39.4 & 38.1 & 48.8   & 58.3 & 51.4 & 53.4 & 53.7  \\
\midrule \rowheader \multicolumn{19}{l}{\textit{\textbf{Classification-Supervised Models}}} \\
DeiT III~\cite{touvron2022deit}         & ViT-B16   & IN-22k      & 21.8 & 12.4 &  9.4 & 16.9   & 41.8 & 31.7 & 34.6 & 36.8   & 37.7 & 33.1 & 35.4 & 35.0   & 17.1 & 15.6 & 16.0 & 16.1  \\
DeiT III~\cite{touvron2022deit}         & ViT-L16   & IN-22k      & 23.6 & 15.0 & 12.3 & 18.5   & 47.0 & 35.7 & 39.2 & 41.4   & 40.6 & 35.3 & 38.1 & 37.5   & 27.7 & 24.8 & 25.1 & 25.4  \\
ConvNeXt~\cite{liu2022convnet}          & CNXT-B    & IN-22k      &  4.6 &  3.8 &  4.1 &  4.2   &  7.9 &  6.6 &  6.6 &  7.5   & 20.7 & 16.9 & 18.6 & 18.8   & 14.5 & 12.5 & 13.3 & 13.5  \\
\midrule \rowheader \multicolumn{19}{l}{\textit{\textbf{Vision Language Models}}} \\
CLIP\cite{radford2019language}          & ViT-B16   & WIT         &  6.1 &  5.4 &  5.7 &  5.7   &  5.8 &  4.3 &  3.7 &  4.9   &  4.9 &  3.6 &  3.5 &  4.3   &  6.0 &  3.5 &  2.8 &  4.7  \\
CLIP\cite{radford2019language}          & ViT-L14   & WIT         &  4.6 &  4.4 &  4.4 &  4.6   &  4.0 &  3.4 &  3.7 &  3.9   &  3.5 &  2.6 &  3.0 &  3.2   &  3.2 &  2.6 &  2.9 &  3.0  \\
CLIP\cite{openclip}                     & ViT-B16   & LAION       &  5.6 &  5.0 &  4.2 &  5.2   &  5.5 &  4.1 &  4.1 &  4.8   &  6.4 &  3.1 &  2.8 &  5.0   &  7.2 &  3.2 &  2.6 &  5.3  \\
CLIP\cite{openclip}                     & CNXT-B    & LAION       &  6.0 &  5.3 &  5.4 &  5.5   &  5.3 &  4.1 &  4.1 &  4.8   & 24.8 & 19.9 & 20.4 & 21.9   & 21.5 & 18.6 & 21.1 & 19.1  \\
CLIP + AugReg\cite{openclip}            & CNXT-B    & LAION       &  5.4 &  4.8 &  4.2 &  5.1   &  6.8 &  5.6 &  5.7 &  6.1   & 27.4 & 22.9 & 22.4 & 24.4   & 27.9 & 24.3 & 27.8 & 25.7  \\
SigLIP~\cite{zhai2023sigmoid}           & ViT-B16   & WebLI       &  8.4 &  7.4 &  8.1 &  8.1   &  8.7 &  7.2 &  7.6 &  8.2   &  7.7 &  6.7 &  6.4 &  7.1   &  5.9 &  5.2 &  4.9 &  5.6  \\
SigLIP~\cite{zhai2023sigmoid}           & ViT-L16   & WebLI       &  7.1 &  5.7 &  5.8 &  6.5   &  7.3 &  6.0 &  6.2 &  6.7   &  6.7 &  5.3 &  5.5 &  6.3   &  4.4 &  3.9 &  4.0 &  4.3  \\
\midrule \rowheader \multicolumn{19}{l}{\textit{\textbf{Text-Conditioned Image Generation Models}}} \\
StableDiffusion~\cite{rombach2022high}  & UNet      & LAION       & 14.2 &  5.7 &  3.8 & 10.1   & 58.0 & 34.6 & 28.5 & 46.4   & 21.9 & 15.5 & 12.2 & 17.8   &  4.6 &  4.1 &  4.2 &  4.2  \\
\midrule \rowheader \multicolumn{19}{l}{\textit{\textbf{Densely-Supervised Models}}} \\
SAM~\cite{kirillov2023sam}              & ViT-B16   & SA-1B       &  9.2 &  5.5 &  4.6 &  7.4   & 16.4 & 10.7 &  8.7 & 13.0   & 29.0 & 18.6 & 14.0 & 23.2   & 30.9 & 19.2 & 14.4 & 24.7  \\
SAM~\cite{kirillov2023sam}              & ViT-L16   & SA-1B       &  9.9 &  6.1 &  5.4 &  8.0   & 22.6 & 15.8 & 12.5 & 18.3   & 34.8 & 23.1 & 17.0 & 28.2   & 30.2 & 18.1 & 13.0 & 24.1  \\
MiDaS~\cite{ranftl2021dpt}              & ViT-L16   & MIX 6       & 15.6 & 10.2 &  8.7 & 13.0   & 27.3 & 22.8 & 23.2 & 24.5   & 28.2 & 23.4 & 25.1 & 25.5   & 25.8 & 21.3 & 23.6 & 23.4  \\
\bottomrule
\end{tabularx}
\end{table*}